# Type-2 Fuzzy Reliability-Redundancy Allocation Problem and its solution using Particle Swarm Optimization Algorithm

**Zubair Ashraf*** **Pranab K. Muhuri*** **Q. M. Danish Lohani+** **Mukul Lata Roy***

Department of Computer Science*, Department of Mathematics+

South Asian University, Akbar Bhavan, Chanakyapuri, New Delhi-110021, India

Email: ashrafzubair786@gmail.com, pranabmuhuri@cs.sau.ac.in, danishlohani@cs.sau.ac.in, mukul.lata@gmail.com

Abstract — In this paper, the fuzzy multi-objective reliability-redundancy allocation problem (FMORRAP) is proposed, which maximizes the system reliability while simultaneously minimizing the system cost under the type-2 fuzzy uncertainty. In the proposed formulation, the higher order uncertainties (such as parametric, manufacturing, environmental and designers' uncertainty) associated with the system are modeled with interval type-2 fuzzy sets (IT2 FS). The footprint of uncertainty of the interval type-2 membership functions (IT2 MFs) accommodates these uncertainties by capturing the multiple opinions from several system experts. We consider IT2 MFs to represent the sub-system reliability and cost, which are to be further aggregated by using extension principle to evaluate the total system reliability and cost according to their configurations, i.e., series-parallel and parallel-series. We proposed a particle swarm optimization (PSO) based novel solution approach to solve the FMORRAP. To demonstrate the applicability of two formulations, namely series-parallel FMORRAP and parallel-series FMORRAP, we performed experimental simulations on various numerical datasets. The decision makers/system experts assign different importance to the objectives (system reliability and cost), and these preferences are represented by sets of weights. The optimal results are obtained from our solution approach, and the Pareto-optimal front is established by using these different weight sets. The genetic algorithm (GA) was implemented to compare the results obtained from our proposed solution approach. A statistical analysis was conducted between PSO and GA, and it was found that the PSO based Pareto solution outperforms the GA.

**Keywords –** Multi-objective reliability allocation problem, Parallel-series and Series-parallel system, Type-2 fuzzy reliability, Type-2 fuzzy cost, Particle swarm optimization.

## 1. Introduction

As the world becomes increasingly tech-savvy, a substantial amount of global expenses goes into the development of very reliable products of a high standard. As a result, the estimation of reliability is considered entirely a useful tool that can be used to assess the lifetime of various goods in different sectors, such as robotics, power plants, satellites, and so on. Since each of these sectors has different needs, they customize procedures for reliability assessment to their specifications. Therefore, this area of work has consistently been the focus of the research community in the last few decades. The reliability of a product has been defined as the ability of that product to perform its intended task under certain operational conditions and for a given time period (Rausand 2014). Now, this product may be a sub-system or an element. The task of the product may be a single operation or a combination of functions, while the period may range from days to years. Decision makers utilize reliability assessment in order to make fundamental decisions regarding the designing of a system. These necessary decisions are meant to handle any number of issues that may arise before or after the system has started its operation. The issues can be anything ranging from implementation, handling of constraints, processing, system maintenance, etc.

The reliability of a system can be increased in two ways: (i) by improving the reliability of the individual components, and (ii) via component redundancy. Due to the unlikelihood of having components of high reliability available at the same time of requirement, the first method is not ideal. The second method involves including several dispensable or redundant components in a system such that, if any individual component were to fail, it would not affect the operation of the rest of the system. However, with the addition of these superfluous components to the system, other factors such as weight, cost, and volume are affected negatively. Because of this, it is essential to find a trade-off between the system's reliability and its redundancies under specific design constraints. This problem is known as the reliability-redundancy allocation problem (RRAP) (Fyffe et al. 1968; Misra 1971; Tillman et al. 1977, Bulfin and Liu 1985).

Chern studied the complexity issues of the RRAP problem and proved that it belonged to the NP-hard class (Chern 1992). Therefore, heuristic or meta-heuristic based solution approaches would achieve optimality rather than traditional solutions methods. The RRAP has been addressed both as a single-objective optimization problem (SOOP) and multi-objective optimization problem (MOOP) with a number of heuristic and meta-heuristic methods (Chern 1992; Kim and Yum 1993; Zia and Coit 2010). The research on the RRAP has been done for different system structures, such as series, parallel, series-parallel, and parallel-series, etc. (Rausand 2014). From all these structures, the series-parallel and parallel-series systems are the most commonly used in the research of system design. Fig. 1 shows typical series-parallel and parallel-series system configurations, with the $m$ sub-systems. This arrangement of the sub-systems and components guarantees that the malfunction of individual components will not cause the whole system to shut down.

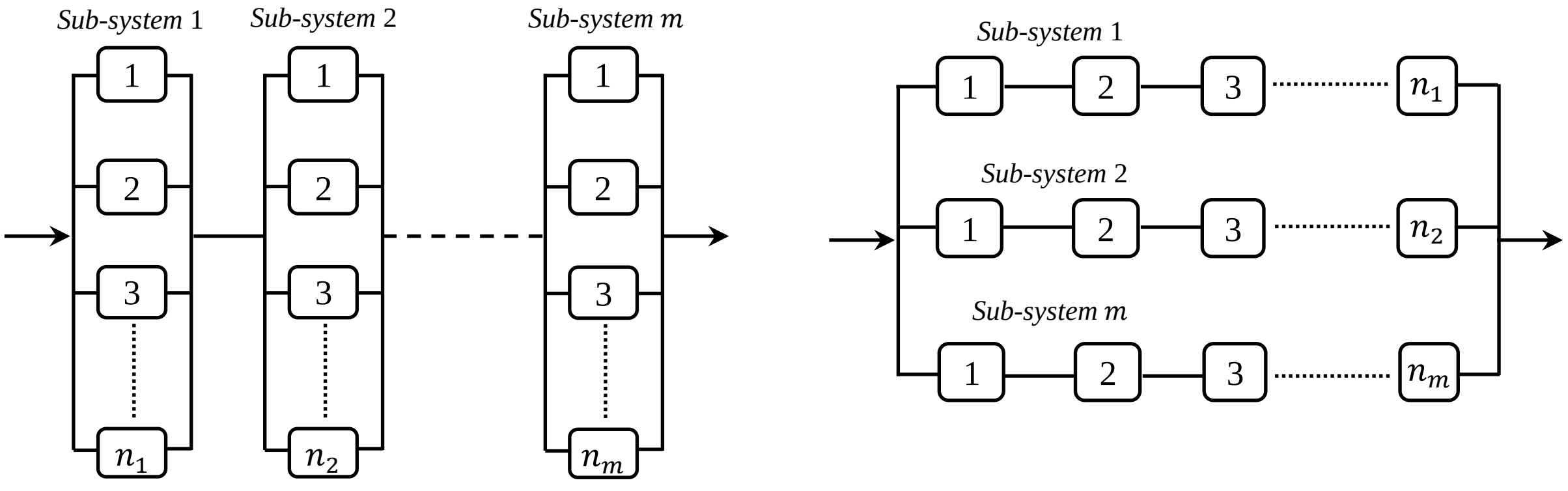

(a) Series-parallel configuration

(b) Parallel-series configuration

**Fig. 1:** Series-parallel and parallel-series systems with $m$ subsystems

The classical RRAP models are formulated under the assumption that all the information regarding the reliability and cost parameters of the system are precise or deterministic. However, when we speak of the design of systems in real life, various uncertainties are found to exist at different stages of the process. These uncertainties are related to important aspects of the system, such as reliability, cost, weights, etc., and can arise for a number of reasons (Nannapaneni and Mahadevan 2016; Zio 2009; Muhuri et al. 2017):

- Lack of information available to the designers (designers' uncertainty).
- Imprecise data due to approximation and randomness (parametric uncertainty).
- Uncertainty associated with the materials of manufactured components (manufacturing uncertainty).
- Unpredictable environment at the time of system deployment (environmental uncertainty).

Incorporating these uncertainties into the design process of the system impacts the overall system development significantly. As such, a great deal of research has been dedicated to the modeling of RRAP while taking these uncertainties into consideration. In particular, fuzzy reliability theory has been incorporated to deal with these uncertainties with great success (Klir and Yuan 2008; Soltani 2014; Garg and Sharma 2013; Ravi et al. 2000; Huang 1997; Mon and Cheng 1994; Mahapatra and Roy 2006). The incentives behind modeling the uncertain sub-system components of the RRAP can be listed as follows:

a) Due to different costs of raw materials as well as different manufacturers of the various sub-system components, there may be different values of cost, weight, volume and reliability for each sub-system.
b) Designers may have very little knowledge of these component parameters, and may incorporate approximate values into their design; since these approximate values are not precise, fuzzy numbers are ideal to model these parameters.
c) The practical interpretation of the uncertainties of the system by the designers differs from individual to individual, which leads to a range of membership values of cost, reliability, etc.
d) Additionally, parameters of the systems such as reliability, cost, and temperature etc. are just predicted by the system experts during the designing phase.

Therefore, type-1 fuzzy sets (T1 FS) are not sufficient to handle the multiple opinions of the decision makers. To model the various uncertainties that come from different sources and the different opinions of system experts, type-2 fuzzy sets (T2 FS) (Zadeh 1975) are the most suitable approach and have already been applied in a number of real-life applications (Bustince et al. 2016a, b; Chen and Lee 2010; Lee and Chen 2008; Castillo et al. 2016; Ashraf et al. 2014, 2015, 2017; Muhuri and Shukla 2017; Valdez et al. 2017; Gaxiola et al. 2016).

In granular computing, the known information about models is integrated to design an enhanced system that is capable of better performance via the incorporation of different levels of specificity based on human perception (Pedrycz and Chen 2011, 2015a, b). The uncertainties involved in the information can be modeled using type-2 fuzzy sets (T2 FSs), which permits the different degrees of membership, in the form of the footprint of uncertainty (FOU), to portray the imprecise and ambiguous parameters of the system (Sanchez et al. 2017).

Based on the above discussion, we suggest handling the reliability and cost of the sub-systems by modeling them with T2 FSs. Due to the higher burden of computations of T2 FSs, we used interval type-2 fuzzy sets (IT2 FSs). Hence, a novel fuzzy multi-objective reliability-redundancy allocation problem (FMORRAP) is proposed, which maximizes the system reliability while simultaneously minimizing the system cost under type-2 fuzzy uncertainty.

The significant contributions of this paper are as follows:

1) A new FMORRAP is proposed to maximize the reliability and minimize the cost of the system under the constraints of weight and volume.
2) The objectives of FMORRAP are modelled as interval type-2 fuzzy sets (IT2 FS) with their corresponding interval type-2 fuzzy membership functions (IT2 MFs).
3) The IT2 MFs of the total system reliability and cost are calculated by aggregating the sub-system reliabilities and costs using Zadeh’s extension principle, according to their configurations, i.e., series-parallel and parallel-series.
4) We proposed a particle swarm optimization (PSO) based novel solution approach to find the optimal solution of the FMORRAP formulations for both series-parallel and parallel-series configurations.
5) Experimental simulations were performed for the two formulations, namely series-parallel FMORRAP and parallel-series FMORRAP, using numerical examples.
6) To demonstrate the applicability of our proposed formulations, we performed experimental simulations on real-life numerical datasets.
7) The Pareto-optimal front was established by using different combinations of weight vectors to explore the optimality of the front.
8) The genetic algorithm (GA) has also been implemented to perform a thorough comparison with our proposed solution approach, and statistical analysis has been conducted to show the efficacy of the optimal results.

The rest of the paper is organized as follows. Section 2 provided the brief overview of the related work done in this field. The mathematical preliminaries that are going to be used in this paper are provided in Section 3. The detailed mathematical formulation of the proposed problem is given in Section 4. Section 5 explains the complete solution approach to solve the problem. In Section 6, we have performed the experimental simulations by considering numerical datasets and demonstrated the efficacy of our proposed approach through comparisons. Finally, we conclude in Section 7.

## 2. Literature Survey

Several surveys were conducted on reliability optimization based on types of problems, structures of systems, and classification of techniques, as found in Tillman et al. (1980), Tzafestas (1980), and Soltani (2014). Also, the works of Kuo and Prasad (2004) and Kuo and Wan (2007) provided an overview of classical reliability optimization along with the advancement in solution approaches. Some works significantly related to this paper are discussed here. Kuo and Prasad (2000) have carried out a brilliant mathematical work on RRAP which has established itself as a base-line in this field and continues to be used as a starting point for multiple further works. In it, the RRAP was elaborated in detail and solution techniques that were developed over the decades were reviewed by the authors. Ramirez-Marquez and Coit (2004) proposed the RRAP model as a multi-state system in which the authors minimized cost under the reliability constraint and solved it heuristically. A method to solve RRAP by using an evolutionary algorithm for a series-parallel configuration was proposed by Tian et al. (2008). Coit and Smith (1996) developed a genetic algorithm (GA) to optimize the linear optimization problem of series-parallel systems. A Monte-Carlo simulation-based PSO method was established by Yeh et al. (2010), to evaluate reliability in conventional series-parallel systems as well as complex network systems. Azadeh et al. (2015) proposed a redundancy-scheduling optimization problem that was solved based on GA for a multi-state series-parallel system for a two-stage manufacturing flow-shop. The MORRAP problem was discussed by Wang et al. (2009) using two objective functions instead of one, i.e., maximizing reliability while simultaneously minimizing the design cost of a system and solved the MORRAP by using a non-dominated sorting genetic algorithm-II (NSGA-II). Khalili-Damghani et al. (2013) proposed a dynamic self-adaptive method based multi-objective particle swarm optimization (MOPSO) to solve the MORRAP. A MOPSO technique was suggested by Dolatshahi-Zand and Khalili-Damghani (2015) to solve the MORRAP, designed for the water resource management control center of supervisory control and data acquisition (SCADA) in Tehran.

Soltani in 2014 presented a number of tables to summarize the literature from the different perspectives of uncertainty modeling in the RRAP (Soltani 2014). An approach for obtaining fuzzy system reliability by $\alpha$-cuts and interval arithmetic was proposed by Cheng and Mon (1993) that involves the use of fuzzy numbers (FNs) over an interval of confidence. The authors also addressed the problem of individual system components having diverse probabilities of failure and suggested the use of fuzzy distributions instead of probability distributions to account for this difference (Mon and Cheng 1994). Further, the authors evaluated the fuzzy system reliability by solving functions of FNs. To deal with the subjectivity surrounding the information obtained from experts, Mahapatra and Roy (2006) solved the multi-objective optimization decision-making for series as well as complex system reliability by using fuzzy sets. In order to analyze the fuzzy reliability of a paper plant's washing system, Komal and Sharma (2014) developed a network and GA based Lambda-Tau technique which hybridized neural network and GA. Chen (1994) represented the reliability of each system as triangular FNs to develop a method for analysis of a fuzzy reliability system that uses FN arithmetic operations.

Huang (1997) proposed a fuzzy multi-objective optimization decision-making method to optimize the reliability of a system with more than one objective. The fuzzy MORRAP problem was developed by Ebrahimipour and Sheikhalishahi (2011) using triangular FNs for the uncertainty in the component parameters. Garg and Sharma (2013) proposed the fuzzy MORRAP with fuzzy objectives and solved it using PSO by utilizing weight preference method. A fuzzy MOOP for complex systems was designed in (Ravi et al. 2000), where not only the reliability but also the cost, weights, and volume of the system were considered as the fuzzy objectives. The authors also considered the influences of various aggregate operators on the solutions of the problem. Recently, Muhuri et al. (2017) proposed a new formulation of the MORRAP, named as interval type-2 fuzzy MORRAP, to model higher order uncertainties in the component parameters of a system using IT2 FNs, and use NSGA-II to obtain the Pareto solutions.

## 3. Preliminaries

**Definition 1** (Zadeh 1965): Fuzzy set (type-1 fuzzy set (T1 FS)) $A$ is characterized by a membership function (MF) of an element over the universal set $X$ as follows:

$$A = \{x, \mu_A(x) | \forall x \in X\} \quad (1)$$

where $\mu_A(x)$ represents the type-1 membership function (T1MF) of $A$, such that $0 \leq \mu_A(x) \leq 1$.

**Definition 2** (John 1998): A type-2 fuzzy set (T2 FS) $\tilde{A}$ is characterized by a type-2 membership function (T2 MF) which itself is fuzzy. $\tilde{A}$ can be expressed as

$$\tilde{A} = \{(x, u), \mu_{\tilde{A}}(x, u) | \forall x \in X, \forall u \in J_x \subseteq [0,1]\} \quad (2)$$

where, $x$ is a primary variable taken from universe of discourse $X$; $u$ is a secondary variable such that $u \in J_x$ at each $x \in X$ and $J_x$ is a primary membership degree of $x$. For indiscrete cases, $\tilde{A}$ is expressed:

$$\tilde{A} = \int_{x \in X} \int_{u \in J_x} \mu_{\tilde{A}}(x, u) / (x, u) \,, \; J_x \subseteq [0,1] \quad (3)$$

**Definition 3** (Mendel and John 2002): When all secondary membership values of T2 FS are 1 i.e. $\mu_{\tilde{A}}(x, u) = 1$, it becomes interval type-2 fuzzy set (IT2 FS).

$$\tilde{A} = \{(x, u), 1 | \forall x \in X, \forall u \in J_x \subseteq [0,1]\} \quad (4)$$

$$\tilde{A} = \int_{x \in X} \int_{u \in J_x} 1/(x, u) \quad , J_x \subseteq [0,1] \quad (5)$$

**Definition 4** (Mendel and Wu 2010): Uncertainty in the primary memberships of an IT2 FS $\tilde{A}$ consists of a bounded region that is called the footprint of uncertainty (FOU). It is the two dimensional support of $\tilde{A}$, that is,

$$FOU(\tilde{A}) = \{(x, u) \in X \times U | \mu_{\tilde{A}}(x, u) > 0\} \quad (6)$$

**Definition 5** (Mendel and Wu 2010): The upper membership function (UMF) and lower membership function (LMF) of $\tilde{A}$ are two T1 MFs that bound FOU ($\tilde{A}$) as shown in Fig 2. The UMF and LMF are associated with the upper bound denoted by $\overline{\mu}_{\tilde{A}}(x)$, $\forall x \in X$ and the lower bound denoted by $\underline{\mu}_{\tilde{A}}(x)$, $\forall x \in X$ of FOU($\tilde{A}$), respectively.

$$\underline{\mu}_{\tilde{A}}(x) = \text{LMF}(\tilde{A}) = \inf\{u | u \in [0, 1], \mu_{\tilde{A}}(x) > 0\} \qquad \forall x \in X \quad (8)$$

$$\overline{\mu}_{\tilde{A}}(x) = \text{UMF}(\tilde{A}) = \sup\{u | u \in [0, 1], \mu_{\tilde{A}}(x) > 0\} \qquad \forall x \in X \quad (9)$$

Note that interval type-2 membership function (IT2 MF) $\mu_{\tilde{A}}(x)$ can be expressed as

$$\mu_{\tilde{A}}(x) = \left[\underline{\mu}_{\tilde{A}}(x), \overline{\mu}_{\tilde{A}}(x)\right] \quad (10)$$

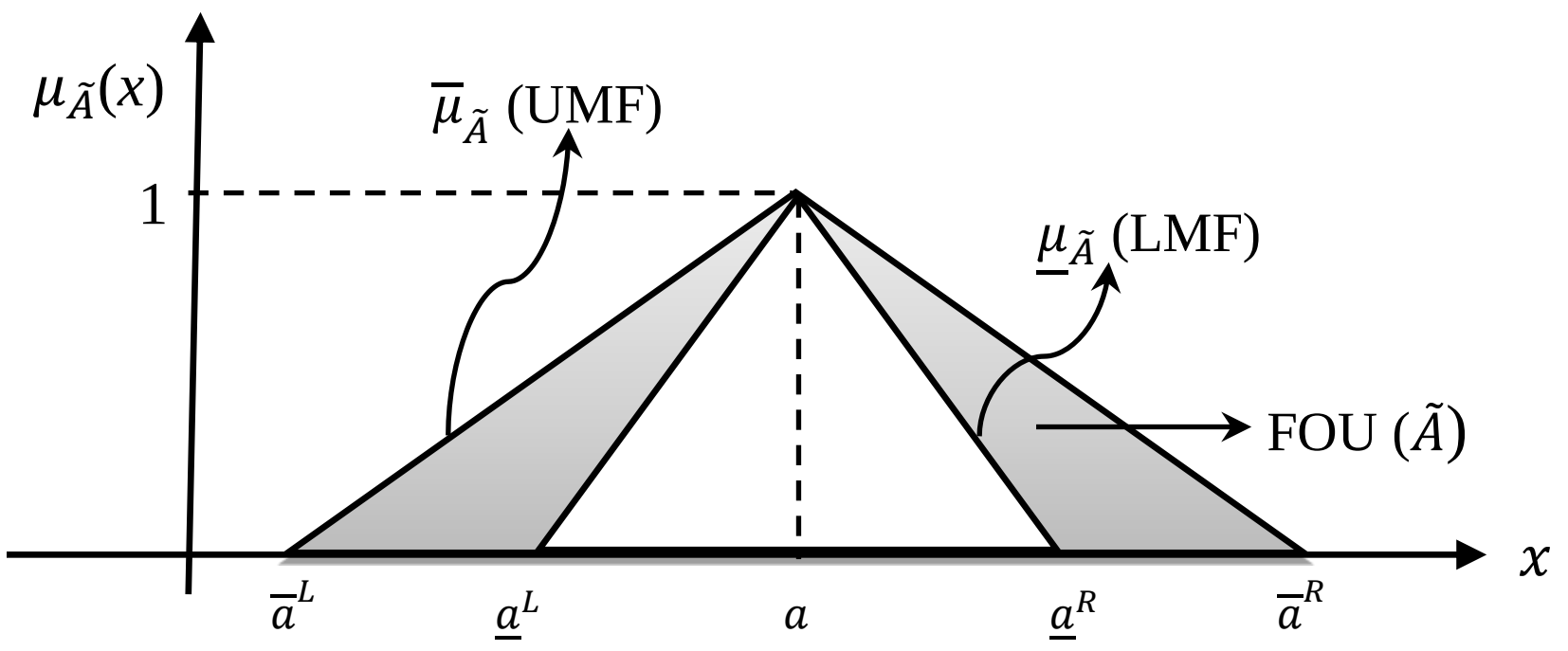

**Fig. 2:** Interval type-2 fuzzy set (IT2 FS).

Fig. 2 graphically represents a triangular IT2 FS in which the LMF and UMF are triangular T1 MFs. In this figure, $\overline{a}^L, \overline{a}^R$ are the left and right ends of UMF and $\underline{a}^L, \underline{a}^R$ are the left and right ends of LMF, which generates the triangular fuzzy number corresponding to $\overline{\mu}_{\tilde{A}}(x; \overline{a}^L, a, \overline{a}^R)$ and $\underline{\mu}_{\tilde{A}}(x; \underline{a}^L, a, \underline{a}^R)$, respectively.

By using the Mendel-John's representation theorem, all the operations on IT2 FSs may be performed using the LMF and UMF, which are T1 FSs. According to this theorem, an IT2 FS $\tilde{A}$ is the union of all T1 FS embedded in FOU ($\tilde{A}$) (John 1998; Karnik and Mendel 2001; Mendel and John 2002; Karnik and Mendel 1998; Mendel and Wu 2010). Therefore, the set theory operations of two IT2 FSs $\tilde{A}$ and $\tilde{B}$ with their corresponding IT2 MFs $\mu_{\tilde{A}}(x)$ and $\mu_{\tilde{B}}(x)$ may be defined as follows:

1) Union $\quad \mu_{\tilde{A} \oplus \tilde{B}}(x) = \mu_{\tilde{A}}(x) \vee \mu_{\tilde{B}}(x) = [\underline{\mu}_{\tilde{A}}(x) \vee \underline{\mu}_{\tilde{B}}(x), \overline{\mu}_{\tilde{A}}(x) \vee \overline{\mu}_{\tilde{B}}(x)]$ (11)

2) Intersection $\quad \mu_{\tilde{A} \otimes \tilde{B}}(x) = \mu_{\tilde{A}}(x) \wedge \mu_{\tilde{B}}(x) = [\underline{\mu}_{\tilde{A}}(x) \wedge \underline{\mu}_{\tilde{B}}(x), \overline{\mu}_{\tilde{A}}(x) \wedge \overline{\mu}_{\tilde{B}}(x)]$ (12)

3) Complement $\quad 1 \ominus \mu_{\tilde{A}}(x) = [1 - \underline{\mu}_{\tilde{A}}(x), 1 - \overline{\mu}_{\tilde{A}}(x)]$ (13)

In Eq. (11)-(12), maximum $t$-conorm and minimum $t$-norm operations are used to perform the union and intersection on two IT2 MFs, respectively.

**Definition 6** (Mahapatra and Roy 2006): Fuzzy multi-objective optimization problem (FMOOP) is defined as follows

$$\begin{aligned} &\text{Maximize/Minimize} \quad && \tilde{f}(x) = \left[\tilde{f}_1(x), \tilde{f}_2(x), \ldots, \tilde{f}_k(x)\right]^T \\ &\text{subject to} && g_j(x) \le or = or \ \ge \ b_j \\ & && x \in X \subseteq \mathbb{R}^n \end{aligned} \qquad (14)$$

**Definition 7** (Mahapatra and Roy 2006): (Complete optimal solution) $x^*$ can be called the complete optimal solution of the FMOOP if and only if $\exists\ x^* \in X$ such that $f_p(x^*) \le f_p(x)\ \forall\ p = 1,2,\ldots,k$ and $\forall\ x \in X$.

When the objective functions of the FMOOP conflict with each other, the complete optimal solution may not exist that causes the concept of Pareto optimality to arise.

**Definition 8** (Mahapatra and Roy 2006): (Pareto optimal solution) $x^*$ can be called a Pareto optimal solution of the FMOOP if and only if there does not exist another $x \in X$ such that $f_p(x^*) \le f_p(x)\ \forall\ p = 1,2,\ldots,k$ and $f_q(x) \ne f_q(x^*)$ for at least one value of $q \in \{1,2,\ldots,k\}$.

**Table 1** Notations used in mathematical model

| Notations | Details | Notations | Details |
|---|---|---|---|
| $m$ | Number of sub-systems | $\tilde{\mathcal{R}}_i$ | Interval Type-2 (IT2) fuzzy reliability of the $i$-th sub-system |
| $n_i$ | Number of redundant components in the $i$-th sub-system | $\tilde{C}_i$ | IT2 fuzzy cost of the $i$-th sub-system |
| $r_i$ | Reliability of each component in the $i$-th sub-system | $\tilde{\mathcal{R}}_s$ | Total IT2 fuzzy reliability function of model |
| $w_i$ | Weight of each component in the $i$-th sub-system | $\tilde{C}_s$ | Total IT2 fuzzy cost function of model |
| $v_i$ | Volume of each component in the $i$-th sub-system | $\underline{\mu}_{\tilde{\mathcal{R}}_S}, \overline{\mu}_{\tilde{\mathcal{R}}_S}$ | Lower membership function (LMF) and upper membership function (UMF) of the $\mu_{\tilde{\mathcal{R}}_S}$ total reliability |
| $\alpha_i, \beta_i$ | Shaping & Scaling factor of each component the $i$-th sub-system | $\underline{\upsilon}_{\tilde{C}_s}, \overline{\upsilon}_{\tilde{C}_s}$ | Lower membership function (LMF) and upper membership function (UMF) of the IT2 MF $\upsilon_{\tilde{C}_s}$ total cost |
| $\mathcal{R}_i$ | Reliability of the $i$-th sub-system | $C$ | Upper limit of system cost |
| $\underline{\mathcal{R}}_i^L, \underline{\mathcal{R}}_i^R$ | Left and right ends of LMF in the $i$-th sub-systems' reliability | $\underline{C}_i^L, C_i^R$ | Left and right ends of LMF in the $i$-th sub-systems' cost |
| $\overline{\mathcal{R}}_i^L, \overline{\mathcal{R}}_i^R$ | Left and right ends of UMF in the $i$-th sub-systems' reliability | $\overline{C}_i^L, \overline{C}_i^R$ | Left and right ends of UMF in the $i$-th sub-systems' cost |
| $C_i$ | Cost of the $i$-th sub-system | $V$ | Upper limit of system volume |
| $\mathcal{R}_S$ | Total system reliability function of the model | $W$ | Upper limit of system weight |
| $C_S$ | Total system cost function of model | $V_S$ | Total system volume function of the model |
| $W_S$ | Total system weight function of model | $T$ | Operating temperature |

## 4. Problem Formulation

The mathematical notations used in the problem formulation are given in Table 1. We have considered a MORRAP on the series-parallel configuration in which $m$ sub-systems are connected in series, with each sub-system consisting of redundant components $(n_i; i = 1, \dots, m)$ that are linked in parallel, whereas in parallel-series systems, each sub-system is connected in parallel and the redundant components are joined in series. The series-parallel/parallel-series configuration, as shown in Fig. 1 (a)-(b), ensures that the entire operability of the sub-system remains unaffected even if any component of a sub-system malfunctions. The objective of the classical MORRAP is to maximize the system's reliability while simultaneously minimizing the cost under the weight and volume constraints of the system. Mathematically, the MORRAP formulation is described as below:

$$\begin{gathered}\text{Maximize} \quad \mathcal{R}_s(r,n) \;\; \& \;\; \text{Minimize} \quad C_s(r,n) \\ \text{subject to} \quad W_s(w,n) \le W,\, V_s(v,n) \le V \\ r = (r_1, \dots, r_m),\, n = (n_1, \dots, n_m)\; r_i \in [0,1] \subset \mathbb{R},\, n_i \subset \mathrm{Z}^+\end{gathered} \tag{15}$$

In Eq. (15), the two objective functions, total reliability and cost of a system are represented by $\mathcal{R}_s(r,n)$ and $C_s(r,n)$, whereas the constraint functions, weight and volume of system, are represented by $W_s(w,n)$ and $V_s(v,n)$, respectively.

There are several mathematical models that were developed to formulate the MORRAP; among them the most popular one was the model formulated by Kuo and Prasad (2000). They formulated the MORRAP under the following restrictions:

a) The inherent properties of components are deterministic.

b) The components have binary operation states (active/inactive) and are non-repairable.

c) In each subsystem, all redundant components are identical.

d) The failures of any components are time-independent and do not damage the subsystem.

For a system structured in a series-parallel/parallel-series configuration consisting of $m$ stages, the total reliability (Kuo and Prasad 2000; Wang et al. 2009) has been expressed in Eq. (16):

$$\mathcal{R}_s(r,n) = \begin{cases} \prod_{i=1}^{m}\left[1 - \prod_{j=1}^{n_i}\left(1 - r_{i,j}\right)\right] & for\ series - parallel\ system \\ 1 - \prod_{i=1}^{m}\left[1 - \prod_{j=1}^{n_i} r_{i,j}\right] & for\ parallel - series\ system \end{cases} \tag{16}$$

where, $r_{i,j}$; $i = 1, \dots, m$; $j = 1, \dots, n_i$ denotes the reliability of the $j$-th component of the $i$-th sub-system $n_i$.

The total cost $C_s(r,n)$ of the system is given in Eq. (17), as follows:

$$C_s(r,n) = \sum_{i=1}^{m}\left[\alpha_i\left(-\frac{T}{\ln(r_i)}\right)^{\beta_i} . \left(n_i + \exp(n_i/4)\right)\right] \tag{17}$$

In Eq. (17), $\alpha_i$ and $\beta_i$ respectively characterize the inherent properties of the system: the shaping factor and the scaling factor; $T$ represents the duration of time for which the component at the $i$-th stage must not fail; all values for $\alpha_i$, $\beta_i$, and $T$ are to be provided by the manufacturers of the components.

Similarly, the total weight $W_s(w,n)$ is given in Eq. (18) in terms of the weights of the individual components $(w_i)$ as well as the redundant components $(n_i)$ for the $m$ stage sub-systems.

$$W_s(w,n) = \sum_{i=1}^{m} w_i . \left(n_i + \exp(n_i/4)\right) \tag{18}$$

In Eq. (17)-(18), the term $\exp(n_i/4)$ represents the extra cost and weight of the internal hardware used in the connecting the components of the sub-systems.

The total volume $V_s(v,n)$ of the system is given in Eq. (19) in terms of the individual volumes $(v_i)$ and redundant components $(n_i)$.

$$V_s(v,n) = \sum_{i=1}^{m} w_i v_i^2 n_i^2 \tag{19}$$

Therefore, using Eq. (16)-(19) in Eq. (15), the MORRAP for a series-parallel/parallel-series system can be defined as follows:

$$\text{Maximize} \quad \mathcal{R}_s(r,n) = \begin{cases} \prod_{i=1}^{m}\left[1-\prod_{j=1}^{n_i}(1-r_{i,j})\right]; series-parallel \\ 1-\prod_{i=1}^{m}\left[1-\prod_{j=1}^{n_i} r_{i,j}\right]; parallel-series \end{cases} \tag{20}$$

$$\text{Minimize} \quad C_s(r,n) = \sum_{i=1}^{m}\left[\alpha_i\left(-\frac{t}{\ln(r_i)}\right)^{\beta_i}.\left(n_i+\exp(n_i/4)\right)\right]$$

$$\text{subject to} \quad W_s(w,n) = \sum_{i=1}^{m} w_i.\left(n_i+\exp(n_i/4)\right) \le W$$

$$V_s(v,n) = \sum_{i=1}^{m} w_i v_i^2 n_i^2 \le V.$$

$$r=(r_1,\dots,r_m), n=(n_1,\dots,n_m).\ r_i \in [0,1] \subset \mathbb{R}, n_i \subset \mathrm{Z}^{+}$$

The MORRAP optimization model formulated in Eq. (20) is designed for the crisp (deterministic) situations, i.e., all the functions of the systems are precise. However, the uncertainties that arise in the system are due to the expert judgment, unpredictable environment, and imprecise parameter values (Bustince, Humberto et al. 2015). Therefore, the tradeoff between maximizing the reliability and minimizing the cost is addressed with IT2 fuzzy MOOP model. The IT2 fuzzy MOOP model is developed by associating a degree of satisfaction to the reliability and cost of each sub-system.

Let us consider $\mathcal{R}_1,\mathcal{R}_2,\dots,\mathcal{R}_m$ and $C_1,C_2,\dots,C_m$ represent the reliabilities and costs of $m$ sub-systems, respectively; $\tilde{\mathcal{R}}_1,\tilde{\mathcal{R}}_2,\dots,\tilde{\mathcal{R}}_m$ denotes the IT2 fuzzy reliabilities and $\tilde{C}_1,\tilde{C}_2,\dots,\tilde{C}_m$ denotes the IT2 fuzzy cost of $m$ sub-systems; and $\mu_{\tilde{\mathcal{R}}_i} = \left[\underline{\mu}_{\tilde{\mathcal{R}}_i},\overline{\mu}_{\tilde{\mathcal{R}}_i}\right]$ and $v_{\tilde{C}_i} = \left[\underline{v}_{\tilde{C}_i},\overline{v}_{\tilde{C}_i}\right]$ give the IT2 MFs corresponding to the reliabilities and costs of each sub-system $(i=1,\dots,m)$. The total system reliability $\mathcal{R}_S$ is characterized by the structure function $\varphi$ such that $\mathcal{R}_S = \varphi(\mathcal{R}_1,\mathcal{R}_2,\dots,\mathcal{R}_m)$ (Chen 1994; Mon and Cheng 1994). Therefore, with the IT2 fuzzy reliabilities $(\tilde{\mathcal{R}}_1,\tilde{\mathcal{R}}_2,\dots,\tilde{\mathcal{R}}_m)$ of the sub-systems, the IT2 fuzzy reliability of system using the structure function is defined as follows:

$$\tilde{\mathcal{R}}_s = \tilde{\varphi}(\tilde{\mathcal{R}}_1,\tilde{\mathcal{R}}_2,\dots,\tilde{\mathcal{R}}_m) \tag{21}$$

For series-parallel system, the $m$ sub-systems are connected in series. Therefore, the IT2 fuzzy reliability of the system is calculated as follows:

$$\tilde{\mathcal{R}}_{s_1}(r,n) = \tilde{\mathcal{R}}_1(r_1,n_1)\otimes\tilde{\mathcal{R}}_2(r_2,n_2)\otimes\dots\otimes\tilde{\mathcal{R}}_m(r_m,n_m) \tag{22}$$

From Eq. (22), the IT2 MF of $\tilde{\mathcal{R}}_{s_1}$, represented by $\mu_{\tilde{\mathcal{R}}_{s_1}}(x) = \left[\underline{\mu}_{\tilde{\mathcal{R}}_{s_1}}(x),\overline{\mu}_{\tilde{\mathcal{R}}_{s_1}}(x)\right]$, may be calculated using the intersection operation on the IT2 MFs of the sub-systems' reliabilities as follows:

$$\mu_{\tilde{\mathcal{R}}_{s_1}}(x) = \mu_{\tilde{\mathcal{R}}_1}(x_1)\otimes\mu_{\tilde{\mathcal{R}}_2}(x_2)\otimes\dots\otimes\mu_{\tilde{\mathcal{R}}_m}(x_m) \tag{23}$$

$$\left[\underline{\mu}_{\tilde{\mathcal{R}}_{s_1}}(x),\overline{\mu}_{\tilde{\mathcal{R}}_{s_1}}(x)\right] = \begin{bmatrix}\underline{\mu}_{\tilde{\mathcal{R}}_1}(x_1)\otimes\underline{\mu}_{\tilde{\mathcal{R}}_2}(x_2)\otimes\dots\otimes\underline{\mu}_{\tilde{\mathcal{R}}_m}(x_m), \\ \overline{\mu}_{\tilde{\mathcal{R}}_1}(x_1)\otimes\overline{\mu}_{\tilde{\mathcal{R}}_2}(x_2)\otimes\dots\otimes\overline{\mu}_{\tilde{\mathcal{R}}_m}(x_m)\end{bmatrix}$$

$$= \left[\bigwedge_{i=1}^{m}\underline{\mu}_{\tilde{\mathcal{R}}_i}(x_i), \bigwedge_{i=1}^{m}\overline{\mu}_{\tilde{\mathcal{R}}_i}(x_i)\right] \tag{24}$$

For parallel-series system, the $m$ sub-systems are connected in parallel. Therefore, the IT2 fuzzy reliability of the system is calculated as follows:

$$\tilde{\mathcal{R}}_{s_2} = 1 \ominus ((1 \ominus \tilde{\mathcal{R}}_1(r_1, n_1)) \otimes (1 \ominus \tilde{\mathcal{R}}_2(r_2, n_2)) \otimes \dots \otimes (1 \ominus \tilde{\mathcal{R}}_m(r_m, n_m))) \tag{25}$$

From the Eq. (25), the IT2 MF of $\tilde{\mathcal{R}}_{s_2}$, denoted by $\mu_{\tilde{\mathcal{R}}_{s_2}}(x) = \left[\underline{\mu}_{\tilde{\mathcal{R}}_{s_2}}(x), \overline{\mu}_{\tilde{\mathcal{R}}_{s_2}}(x)\right]$, may be calculated using the intersection and negation operations on the IT2 MFs of the sub-systems' reliabilities as follows:

$$\mu_{\tilde{\mathcal{R}}_{s_2}}(x) = 1 \ominus \left((1 \ominus \mu_{\tilde{\mathcal{R}}_1}(x_1)) \otimes (1 \ominus \mu_{\tilde{\mathcal{R}}_2}(x_2)) \otimes \dots \otimes (1 \ominus \mu_{\tilde{\mathcal{R}}_m}(x_m)\right) \tag{26}$$

$$\left[\underline{\mu}_{\tilde{\mathcal{R}}_{s_2}}(x), \overline{\mu}_{\tilde{\mathcal{R}}_{s_2}}(x)\right] = \begin{bmatrix} 1 \ominus \left(\left(1 \ominus \underline{\mu}_{\tilde{\mathcal{R}}_1}(x_1)\right) \otimes \left(1 \ominus \underline{\mu}_{\tilde{\mathcal{R}}_2}(x_2)\right) \otimes \dots \otimes \left(1 \ominus \underline{\mu}_{\tilde{\mathcal{R}}_m}(x_m)\right)\right), \\ 1 \ominus \left((1 \ominus \overline{\mu}_{\tilde{\mathcal{R}}_1}(x_1)) \otimes (1 \ominus \overline{\mu}_{\tilde{\mathcal{R}}_2}(x_2)) \otimes \dots \otimes (1 \ominus \overline{\mu}_{\tilde{\mathcal{R}}_m}(x_m))\right) \end{bmatrix}$$

$$= \left[1 \ominus \bigwedge_{i=1}^{m} \left(1 \ominus \underline{\mu}_{\tilde{\mathcal{R}}_i}(x_i)\right), 1 \ominus \bigwedge_{i=1}^{m} \left(1 \ominus \overline{\mu}_{\tilde{\mathcal{R}}_i}(x_i)\right)\right] \tag{27}$$

The total IT2 fuzzy cost of the system $\tilde{\mathcal{C}}_S$ corresponding to series-parallel/parallel-series combination is calculated as follows:

$$\tilde{\mathcal{C}}_S(r, n) = \tilde{\mathcal{C}}_1(r_1, n_1) \oplus \tilde{\mathcal{C}}_2(r_2, n_2) \oplus \dots \oplus \tilde{\mathcal{C}}_m(r_m, n_m) \tag{28}$$

From Eq. (28), the IT2 MF of $\tilde{\mathcal{C}}_s$, represented by $\upsilon_{\tilde{\mathcal{C}}_s}(x) = \left[\underline{\upsilon}_{\tilde{\mathcal{C}}_s}(x), \overline{\upsilon}_{\tilde{\mathcal{C}}_s}(x)\right]$, may be calculated using the union operation on the IT2 MFs of the sub-systems' costs as follows:

$$\left[\underline{\upsilon}_{\tilde{\mathcal{C}}_s}(x), \overline{\upsilon}_{\tilde{\mathcal{C}}_s}(x)\right] = \begin{bmatrix} \underline{\upsilon}_{\tilde{\mathcal{C}}_1}(x_1) \oplus \underline{\upsilon}_{\tilde{\mathcal{C}}_2}(x_2) \oplus \dots \oplus \underline{\upsilon}_{\tilde{\mathcal{C}}_m}(x_m), \\ \overline{\upsilon}_{\tilde{\mathcal{C}}_1}(x_1) \oplus \overline{\upsilon}_{\tilde{\mathcal{C}}_2}(x_2) \oplus \dots \oplus \overline{\upsilon}_{\tilde{\mathcal{C}}_m}(x_m) \end{bmatrix}$$

$$= \left[\bigvee_{i=1}^{m} \underline{\upsilon}_{\tilde{\mathcal{C}}_i}(x_i), \bigvee_{i=1}^{m} \overline{\upsilon}_{\tilde{\mathcal{C}}_i}(x_i)\right] \tag{29}$$

Finally, we establish the fuzzy multi-objective reliability-redundancy allocation problem (FMORRAP) model for $m$-state series-parallel/parallel-series systems as follows:

$$\text{Maximize} \quad \widetilde{\mathcal{R}_{s_t}}(r, n) = \{\widetilde{\mathcal{R}_1}(r_1, n_1), \widetilde{\mathcal{R}_2}(r_2, n_2), \widetilde{\mathcal{R}_3}(r_3, n_3), \dots, \widetilde{\mathcal{R}_m}(r_m, n_m)\} \tag{30}$$

$$\text{Minimize} \quad \widetilde{\mathcal{C}_s}(r, n) = \{\widetilde{\mathcal{C}_1}(r_1, n_1), \widetilde{\mathcal{C}_2}(r_2, n_2), \widetilde{\mathcal{C}_3}(r_3, n_3), \dots, \widetilde{\mathcal{C}_m}(r_m, n_m)\}$$

$$\text{Subject to} \quad W_s(r, n) = \textstyle\sum_{i=1}^{m} w_i.(n_i + \exp(n_i/4)) \le W.$$

$$V_s(r, n) = \textstyle\sum_{i=1}^{m} w_i v_i^2 n_i^2 \le V.$$

$$r_{i,min} \le r_i \le r_{i,max}, n_{i,min} \le n_i \le n_{i,max}, r_i \in [0,1] \subset \mathbb{R}, n_i \subset \mathrm{Z}^+; i = 1,2, \dots, m, \mathcal{R}_{s,min} \le \mathcal{R}_s \le \mathcal{R}_{s,max},$$

$$\mathcal{C}_{s,min} \le \mathcal{C}_s \le \mathcal{C}_{s,max}, \mu_{\tilde{\mathcal{R}}_{s_t}}(x) = \left[\underline{\mu}_{\tilde{\mathcal{R}}_{s_t}}(x), \overline{\mu}_{\tilde{\mathcal{R}}_{s_t}}(x)\right]; t \in 1,2 \text{ and } \upsilon_{\tilde{\mathcal{C}}_s}(x) = \left[\underline{\upsilon}_{\tilde{\mathcal{C}}_s}(x), \overline{\upsilon}_{\tilde{\mathcal{C}}_s}(x)\right].$$

where $t = 1$ indicates the series-parallel system and $t = 2$ indicates the parallel-series system; $\mathcal{R}_{s,min}$ and $\mathcal{R}_{s,max}$ are the upper and lower limits of the system reliability; and $\mathcal{C}_{s,min}$ and $\mathcal{C}_{s,max}$ are the upper and lower limits of the system cost.

## 5. Solution Approach

This section elaborates the solution algorithm used to solve the FMORRAP, formulated in Eq. (30). The steps involved in the proposed solution approach are represented in the form of a flowchart in Fig. 3. In our solution approach, we have used a popular meta-heuristic algorithm of optimization known as the particle swarm optimization (PSO), proposed by Kennedy (2011). PSO replicates the social behavior of bird flocks or fish schools in search of food. It is a population-based search algorithm which discovers patterns of birds in a flock for synchronous and rapid movements within a swarm (Engelbrecht 2007; Clerc and Kennedy 2002). Since its original conception, a number of PSO variants were developed and used in a number of applications, e. g., (Clerc 2006; Poli et al. 2007; Olivas et al. 2016; Valdez et al. 2017; Olivas et al. 2017). Due to the fact that the FMORRAP is a non-linear MOOP, proved as a NP-hard problem in (Chern 1992), evolutionary approaches fit better for its solutions. Therefore, PSO is one of the suitable evolutionary optimization algorithms that can be used to find the optimal component reliability $r_i$ and the number of redundant components $n_i$ corresponding to the objective functions. Additionally, we have used type-reduction and defuzzification procedures in our proposed solution approach because the objective functions of the FMORRAP are modeled with IT2 MFs. These IT2 MFs are converted into their respective crisp values during the optimization process of PSO. The complete implementation procedure of our solution approach is given in Algorithm 1.The details of each step involved in the proposed solution algorithm are discussed in the following sub-sections.

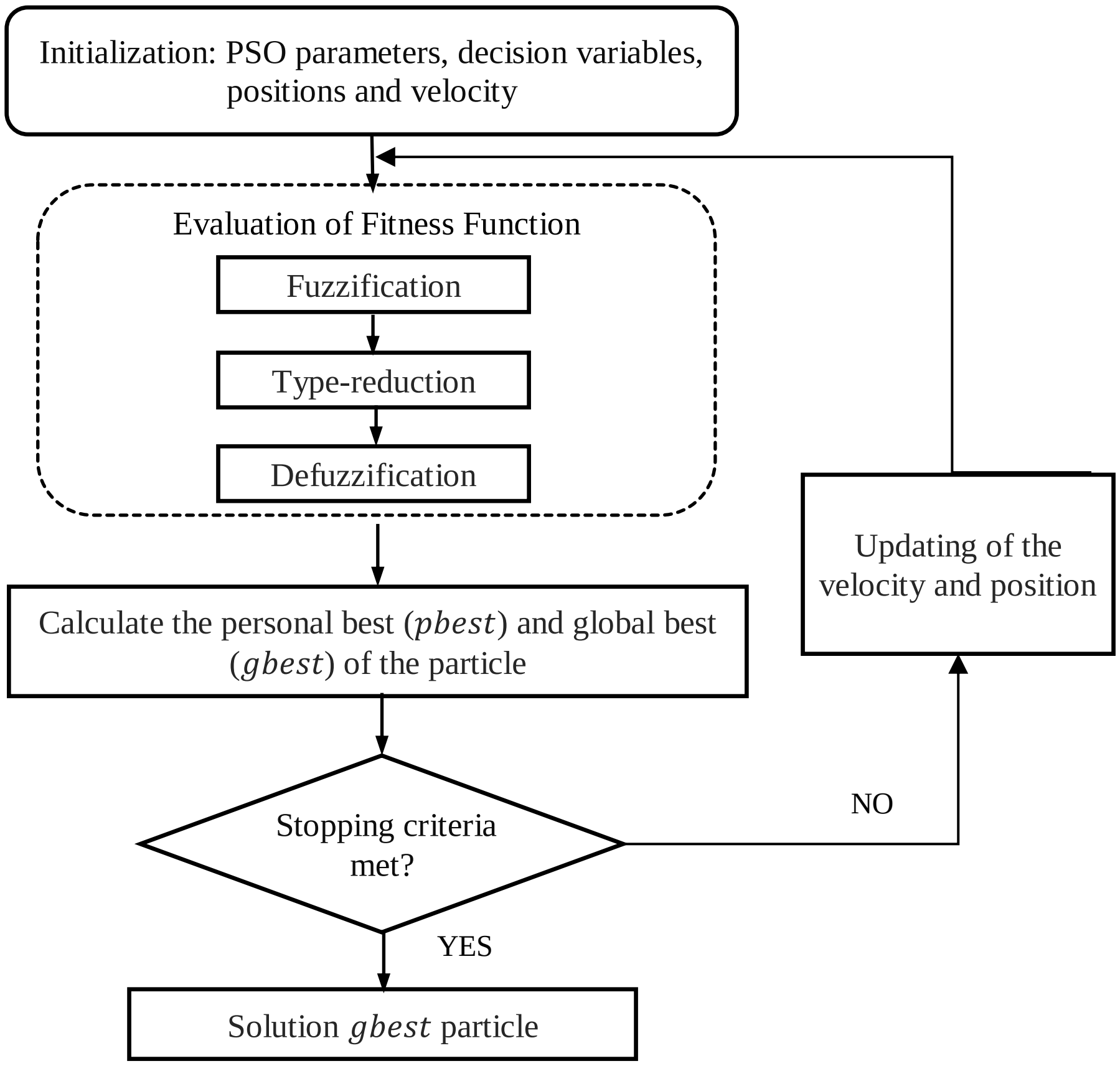

**Fig. 3:** Flow diagram of proposed solution approach

**Algorthim 1: Solution Algorthim**

**Input:** The parameters of PSO and input values of problem
**Output:** Global best particle ($gbest$) and optimal results

Initialization
Fitness function evaluation
    IT2 fuzzification \\ Reliability and cost
    Type-reduction \\ EKM algorithm
    Defuzzification \\ Centroid method
**While** stopping criteria
    Calculate $gbest$ and $pbest$ of each population
    Update the position and velocity of each particle
    Fitness function evaluation
**End**
Global best particle $gbest$ and optimal results

### 5.1. Initialization procedure

In the initialization procedure, (i) the total number of particles in the population ($N_p$) and dimension of each particle ($dim$), (ii) the initial position and velocity of each particle, corresponding to each decision variable ($r, n$), and (iii) the parameters (constriction coefficient set) involved in PSO are assigned initial values. Algorithm 2 gives the randomly initialized positions ($X$) and velocities ($V$) corresponding to each decision variable, and the personal best ($pbest$), and global best ($gbest$) of the total population.

**Algorithm 2: Initialization**

**Input:** Number of particles $N_p$, dimensions of each particles $dim$.
**Output:** Positions ($X$), velocities ($V$), personal best ($pbest$), and global best ($gbest$).

**For** each particle $i = 1 \ to \ N_p$
    **For** each dimension $d = 1 \ to \ dim$ // Initialize all particles' position and velocity
        Evaluate the positions $X_{i,d} \leftarrow Rand(X_{min}, X_{max})$
        Evaluate the velocities $V_{i,d} \leftarrow Rand(V_{min}, V_{max})$
    **end For**
    $pbest \leftarrow X_i$ // Initialize particles' personal best position
    $gbest \leftarrow \emptyset$ // Initialize particles' global best position
**end For**

### 5.2. Fitness function evaluation

In this procedure, the objective functions of FMORRAP, IT2 fuzzy reliability and IT2 fuzzy cost, are evaluated. It involves three steps: (i) fuzzification, (ii) objective evaluation, and (iii) type-reduction and defuzzification. The fuzzification process assigns IT2 MFs to the reliability and cost of the sub-systems, which are used in the evaluation of the objectives (as we proposed in Section 4). Further, the type-reduction and defuzzification procedures provide the defuzzified objective function values of the system.

#### 5.2.1 Fuzzification procedure

The fuzzification procedure captures the uncertainty in differing opinions of multiple experts regarding the fuzzy parameters. The experts provide their opinions about the reliability and cost of the sub-systems in the form of MFs that tend to vary but stay within a certain range. So, we can model reliability and cost of the sub-systems with IT2 MFs. In order to choose a suitable kind of IT2 MF from many (trapezoidal, Gaussian, triangular, etc.), there are certain deductions (Muhuri et al. 2017) that need to be made as follows:

- The reliability of a system is defined as the probability of that system being able to function properly for a certain time period ($T$) without failure. The manufacturers can guarantee this time ($T$), which means that the value of sub-system reliability $\mathcal{R}(T)$ of that time $T$ is unambiguous.
- The sub-system reliability $\mathcal{R}(T)$ at that specific time $T$ should have a membership value of 1. This also implies that there must be only one peak value, and it guarantees that the IT2 MF assigned to the reliability of sub-systems will be normal.
- Since it is also not possible to create a sub-system that works for an infinite amount of time, it is not useful to consider a long tail beyond a certain point in the IT2 MF.
- Therefore, we can rule out Gaussian IT2 MF and trapezoidal IT2 MFs as they are not ideal for modelling the uncertainty in the reliability of the sub-systems.

Hence, an IT2 triangular MF (IT2 TMF) is the best choice for representing the IT2 fuzzy reliability and cost of the sub-systems in FMORRAP. The IT2 TMF $\mu_{\tilde{\mathcal{R}}_i} = \left[\underline{\mu}_{\tilde{\mathcal{R}}_i}, \overline{\mu}_{\tilde{\mathcal{R}}_i}\right]$ for the IT2 fuzzy reliability of the $i$-th sub-system may be defined as follows:

$$\underline{\mu}_{\tilde{\mathcal{R}}_i}\left(x_i; \underline{\mathcal{R}}_i^L, \mathcal{R}_i, \underline{\mathcal{R}}_i^R\right) = \begin{cases} 0, & x_i \le \underline{\mathcal{R}}_i^L \\ \frac{x_i - \underline{\mathcal{R}}_i^L}{\mathcal{R}_i - \underline{\mathcal{R}}_i^L}, & \underline{\mathcal{R}}_i^L \le x_i \le \mathcal{R}_i \\ \frac{\underline{\mathcal{R}}_i^R - x_i}{\underline{\mathcal{R}}_i^R - \mathcal{R}_i}, & \mathcal{R}_i \le x_i \le \underline{\mathcal{R}}_i^R \\ 0, & x_i \le \underline{\mathcal{R}}_i^R \end{cases}; \quad \overline{\mu}_{\tilde{\mathcal{R}}_i}\left(x_i; \overline{\mathcal{R}}_i^L, \mathcal{R}_i, \overline{\mathcal{R}}_i^R\right) = \begin{cases} 0, & x_i \le \overline{\mathcal{R}}_i^L \\ \frac{x_i - \overline{\mathcal{R}}_i^L}{\mathcal{R}_i - \overline{\mathcal{R}}_i^L}, & \overline{\mathcal{R}}_i^L \le x_i \le \mathcal{R}_i \\ \frac{\overline{\mathcal{R}}_i^R - x_i}{\overline{\mathcal{R}}_i^R - \mathcal{R}_i}, & \mathcal{R}_i \le x_i \le \overline{\mathcal{R}}_i^R \\ 0, & x_i \le \overline{\mathcal{R}}_i^R \end{cases} \quad (31)$$

In Eq. (31), $\underline{\mathcal{R}}_i^L, \underline{\mathcal{R}}_i^R$ are the left and right ends of LMF and $\overline{\mathcal{R}}_i^L, \overline{\mathcal{R}}_i^R$ are the left and right ends of UMF, which generate the triangular fuzzy number corresponding to $\underline{\mu}_{\tilde{\mathcal{R}}_i}\left(x_i; \underline{\mathcal{R}}_i^L, \mathcal{R}_i, \underline{\mathcal{R}}_i^R\right)$ and $\overline{\mu}_{\tilde{\mathcal{R}}_i}\left(x_i; \overline{\mathcal{R}}_i^L, \mathcal{R}_i, \overline{\mathcal{R}}_i^R\right)$, respectively. Algorithm 3 gives the procedure of generating the IT2 TMF and different elements of IT2 MF generated using Algorithm 3 are shown in Fig. 4. A typical example of five different sub-system reliabilities modeled with IT2 MFs, which were generated using the Algorithm 3, is shown in Fig. 5. Similarly, we can also generate the IT2 TMF $\nu_{\tilde{C}_i} = \left[\underline{\nu}_{\tilde{C}_i}, \overline{\nu}_{\tilde{C}_i}\right]$ for the IT2 fuzzy cost of the $i$-th sub-system.

### 5.2.2 Objective evaluation

The IT2 TMFs $\mu_{\tilde{\mathcal{R}}_i}$ and $\nu_{\tilde{C}_i}$; $i = 1, \dots, m$, of reliability and cost of the sub-systems are used to calculate the total IT2 fuzzy system reliability $\tilde{\mathcal{R}}_S$ of series-parallel and parallel-series system (as given in Eq. (24) and Eq. (27)) and IT2 fuzzy system cost $\tilde{C}_S$ (as given in Eq. (29)), respectively.

**Algorithm 3:** Generation of IT2 TMF

**Input:** Given $\mathcal{R}_i$, $\mathcal{R}_i^L$ and $\mathcal{R}_i^R$ such that $\mathcal{R}_i^L \le \mathcal{R}_i \le \mathcal{R}_i^R$ and $\mathcal{R}_i \in [a, b]$.

**Output:** IT2 MF $\mu_{\tilde{\mathcal{R}}_i}(x) = \left[\underline{\mu}_{\mathcal{R}_i}(x), \overline{\mu}_{\mathcal{R}_i}(x)\right]$.

**1:** Evaluate the lower $\underline{\mathcal{R}}_i^L$ and upper $\overline{\mathcal{R}}_i^L$ ends of $\mathcal{R}_i^L$ as follows:

$$\underline{\mathcal{R}}_i^L = \mathcal{R}_i^L + (\mathcal{R}_i - \mathcal{R}_i^L) * rand; \quad \overline{\mathcal{R}}_i^L = \mathcal{R}_i^L - (\mathcal{R}_i^L - a) * rand$$

**2:** Evaluate the lower $\underline{\mathcal{R}}_i^R$ and upper $\overline{\mathcal{R}}_i^R$ ends of $\mathcal{R}_i^R$ as follows:

$$\underline{\mathcal{R}}_i^R = \mathcal{R}_{\mathrm{i}}^R - (\mathcal{R}_{\mathrm{i}}^R - \mathcal{R}_i) * rand; \quad \overline{\mathcal{R}}_i^R = \mathcal{R}_{\mathrm{i}}^R + (b - \mathcal{R}_{\mathrm{i}}^R) * rand$$

**3:** Calculate the LMF $\underline{\mu}_{\mathcal{R}_i}\left(x; \underline{\mathcal{R}}_i^L, \mathcal{R}_{\mathrm{i}}^{\mathrm{L}}, \underline{\mathcal{R}}_i^L\right)$ of IT2 MF using Eq. (31).

**4:** Calculate the UMF $\overline{\mu}_{\mathcal{R}_i}\left(x; \overline{\mathcal{R}}^L, \mathcal{R}, \overline{\mathcal{R}}^R\right)$ of IT2 MF using Eq. (31).

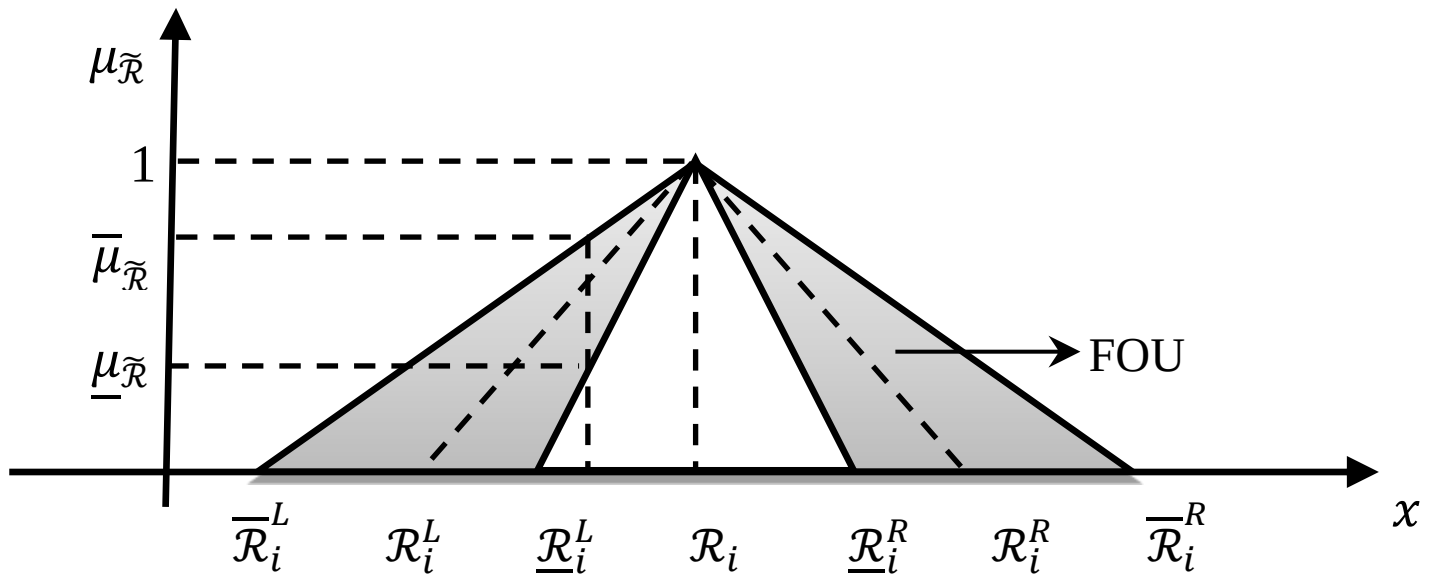

**Fig. 4:** The representation of parameters and IT2TMF using Algorithm 3.

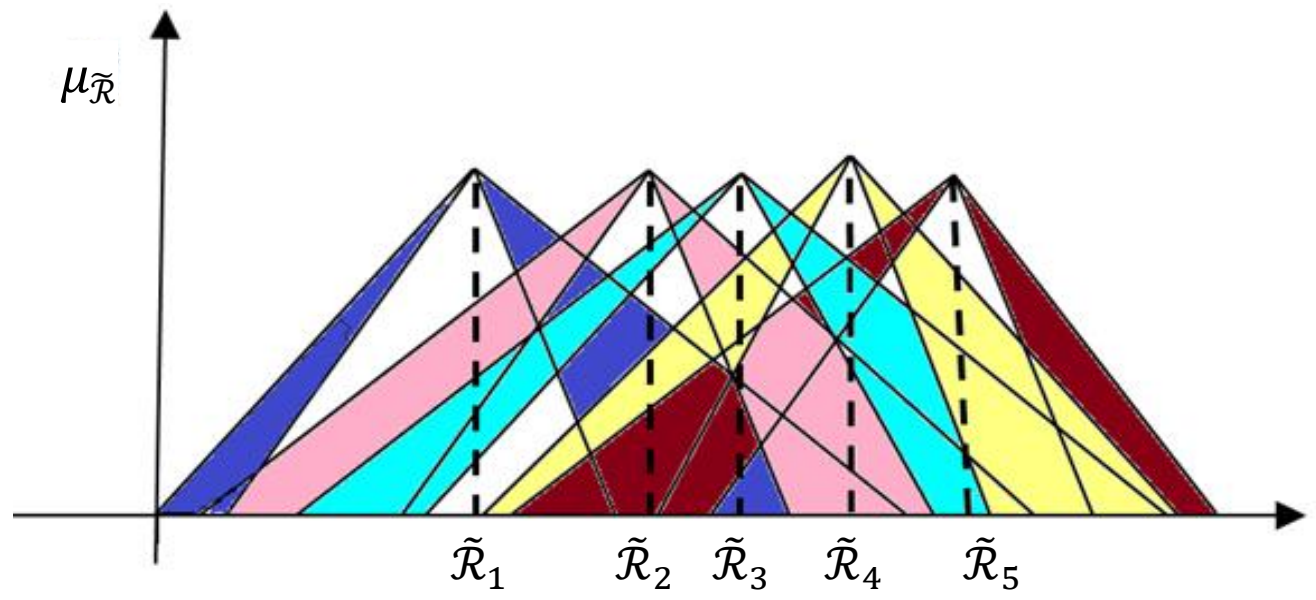

**Fig. 5:** IT2 fuzzy reliabilities for five sub-systems generated by Algorithm 3 (Muhuri et al. 2017).

**Algorithm 4: EKM Algorithm**

| Algorithm for $c_l(\tilde{A})$ | Algorithm for $c_r(\tilde{A})$ |
|---|---|
| $c_l(\tilde{A}) = \min_{\forall\theta_i\in[\underline{\mu},\bar{\mu}]} \left( \sum_{i=1}^{N} x_i\theta_i \Big/ \sum_{i=1}^{N} \theta_i \right)$ | $c_r(\tilde{A}) = \max_{\forall\theta_i\in[\underline{\mu},\bar{\mu}]} \left( \sum_{i=1}^{N} x_i\theta_i \Big/ \sum_{i=1}^{N} \theta_i \right)$ |
| 1. Set $k = [N/2.4]$ (the nearest integer to $N/2.4$) and compute: $a = \sum_{i=1}^{k} x_i \bar{\mu}_{\tilde{A}}(x_i) + \sum_{i=k+1}^{k} x_i \underline{\mu}_{\tilde{A}}(x_i)$; $b = \sum_{i=1}^{k} \bar{\mu}_{\tilde{A}}(x_i) + \sum_{i=k+1}^{k} \underline{\mu}_{\tilde{A}}(x_i)$ | Set $k = [N/1.7]$ (the nearest integer to $N/1.7$) and compute: $a = \sum_{i=1}^{k} x_i \underline{\mu}_{\tilde{A}}(x_i) + \sum_{i=k+1}^{k} x_i \bar{\mu}_{\tilde{A}}(x_i)$; $b = \sum_{i=1}^{k} \underline{\mu}_{\tilde{A}}(x_i) + \sum_{i=k+1}^{k} \bar{\mu}_{\tilde{A}}(x_i)$ |
| 2. Compute $c' = a/b$ | |
| 3. Find $k' \in [1, N-1]$ such that $x_{k'} \le c' x_{k'+1}$ | |
| 4. Check if $k' = k$. If yes, stop and set $c' = c_l(L)$, and $k = L$. If no, go to step 5. | Check if $k' = k$. If yes, stop and set $c' = c_r(R)$, and $k = R$. If no, go to step 5. |
| 5. Compute$s = \text{sign}(k' - k)$ and $a' = a + s \sum_{i=\min(k,k')+1}^{\max(k,k')} x_i \left[\bar{\mu}_{\tilde{A}}(x_i) - \underline{\mu}_{\tilde{A}}(x_i)\right]$; $b' = b + s \sum_{i=\min(k,k')+1}^{\max(k,k')} \left[\bar{\mu}_{\tilde{A}}(x_i) - \underline{\mu}_{\tilde{A}}(x_i)\right]$ | Compute$s = \text{sign}(k' - k)$and $a' = a - s \sum_{i=\min(k,k')+1}^{\max(k,k')} x_i \left[\bar{\mu}_{\tilde{A}}(x_i) - \underline{\mu}_{\tilde{A}}(x_i)\right]$; $b' = b - s \sum_{i=\min(k,k')+1}^{\max(k,k')} \left[\bar{\mu}_{\tilde{A}}(x_i) - \underline{\mu}_{\tilde{A}}(x_i)\right]$ |
| 6. Compute $c''(k') = a'/b'$ | |
| 7. Set $c' = c''(k'), a = a', b = b'$and go to Step 2. | |

#### 5.2.3 Type-reduction and defuzzification procedures

The outputs of the optimization problem are crisp values of total reliability and cost of the system, so the process of type-reduction and defuzzification transforms the IT2 MFs into crisp values. Here, the popular Enhanced Karnik and Mendel (EKM) algorithm (Mendel and Wu 2010) has been used for type-reduction, and the centroid method (Mendel 2007; Mendel and Liu 2013) has been used for defuzzification. The centroid $C_{\tilde{A}}(x)$ of an IT2 FS $\tilde{A}$ is the union of the centroids of all its embedded T1 FSs. That is,

$$C_{\tilde{A}}(x) = \{c_l(\tilde{A}), \dots, c_r(\tilde{A})\} = [c_l(\tilde{A}), c_r(\tilde{A})] \tag{32}$$

The complete procedure of the EKM algorithm, to calculate the left end $c_l$ and right end $c_r$ points of the interval, may be found in (Mendel and Wu 2010; Mendel and Liu 2013). We have included the salient steps of the EKM algorithm in Algorithm 4. The procedure of defuzzification by centroid method uses left centroid ($c_l$) and right centroid ($c_r$). The final crisp output ($y_d$) is the simple mean of left centroid and right centroid:

$$y_d = \frac{c_l + c_r}{2} \tag{33}$$

### 5.3. Finding $gbest$ and $pbest$ of population

The particles of the population move within the search space by the use of a simple mathematical formula over velocity and position at each instance. Each particle's movements are biased by its previous personal best ($pbest$) or local best position. Simultaneously, each particle is led towards the overall best known position of the space, known as the global best ($gbest$). Both $pbest$ and $gbest$ help in moving the swarm towards the best possible solution. Algorithm 5 gives the procedure for finding $pbest$ and $gbest$ in the population.

**Algorithm 5: Finding the $gbest$ and $pbest$**

**Input:** Objective function $f$, number of particles $N_p$, dimension of particles $dim$.
**Output:** Global best particle $gbest$ and personal best particles $pbest$.

```
1:  While maximum iterations or stopping criteria is not attained
2:      For each particle i = 1 to N_p
3:          If f(X_i) < f(pbest_i)
4:              Update the personal best pbest_i ← X_i
5:          End If
6:          If f(pbest_i) < f(gbest)
7:              Update the global best gbest ← pbest_i
8:          End If
9:      End For
10: End While
```

### 5.4. Updating the position and velocity of each particle

If $t$ denotes the time instance, the movement of the $i$-th particle at the instance $t+1$ is $X_i(t+1)$, calculated by adding its previous position $X_i(t)$ to its velocity $V_i(t)$ at the instance $t$, as follows:

$$X_i(t+1) = X_i(t) + V_i(t) \tag{34}$$

Further, the velocity $V_i(t+1)$ of the $i$-th particle at the instance $t+1$ is updated using the previous velocity $V_i(t)$, previous position $X_i(t)$, $pbest_i(t)$, and $gbest(t)$, as follows:

$$V_i(t+1) = \chi * [V_i(t) + c_1 \times r_1 \times (pbest_i(t) - X_i(t)) + c_2 \times r_2 \times (gbest(t) - X_i(t))] \tag{35}$$

In Eq. (35), $c_1$, $c_2$ and $k$ represent the construction coefficient set of the PSO and $\chi$ is the construction coefficient calculated as $\chi = \frac{2*k}{|2-\phi-\sqrt{\phi(\phi-4)}|}$, $k \in [0,1]$, $\phi_1 = c_1 r_1$, $\phi_2 = c_2 r_2$, and $\phi = \phi_1 + \phi_2$. $r_1$ & $r_2$ are random variables with uniform distribution between 0 and 1 (Clerc 2006). The steps required for updating the position and velocity of the particles in the PSO algorithm are given in Algorithm 6.

**Algorithm 6: Updating the position and velocity of each particle**

**Input:** Number of particles $N_p$, dimension of particles $dim$, velocity $V$, position $X$.
**Output:** Updated velocity $V'$, updated position $X'$.

```
1:  For each particle i = 1 to N_p
2:      For each dimension d = 1 to dim
3:          Update the velocity V'_{i,d} ← (V_{i,d}, pbest_i, gbest)     // Using Eq. (35)
4:          Update the position X'_{i,d} ← (X_{i,d}, V_{i,d})            // Using Eq. (34)
5:      end For
6:  end For
```

### 5.5. Constraint handling

For solving a constrained optimization problem, handling the constraints associated with the objective functions is important. Finding the feasible solution to an optimization problem, whether it is single or multi-objective, can become very difficult in the presence of various kinds of inequalities. For handling such situations, many methods have been proposed. Of these methods, the most suitable one is the method using penalty functions (Coello Coello 2002; Marler and Arora 2010). In our solution approach, we have used the penalty guided based function in order to manage the constraints in PSO. This is done by using a modified function $J$ to solve a problem in the search space $X$ as follows:

$$J(x) = \begin{cases} F + \sum_i g_i(x), & if \quad x \notin X \\ F(x), & if \quad x \in X \end{cases} \tag{36}$$

Where $F$ represents the functional values of the worst feasible solution $x$ and $g_i$ is the constraint function.

## 6. Experimental Simulations and Comparisons

In this section, we have used the proposed solution approach in order to solve the FMORRAP for the two configuration systems, i.e., series-parallel and parallel-series. Both these FMORRAPs are solved with the proposed PSO based solution approach by using a traditional adaptive weighted-sum method (Kim and de Weck 2005; Marler and Arora 2010). Corresponding to each objective, the weight vectors are provided by system experts to compute the IT2 fuzzy objective for optimization. The IT2 MFs of the objective functions make up the IT2 fuzzy region of satisfaction (as the objective), which is constructed by system experts. Further, suitable input parameters of the system are used to find the two objectives (reliability and cost) for different combinations of weights. We have also implemented real-parameter Genetic Algorithm (GA) proposed by Golberg (1989) for comparison. The detailed explanation of GA can be found in (Deb 2001).

To verify that the two competing algorithms, PSO and GA, have a significant difference in their performance, we have conducted different runs and performed statistical analysis. In the statistical sense, the mean, standard deviation (SD), median, $t$-test and multivariate analysis of variance (M-ANOVA) are evaluated to show the significance of the FMORRAP with the randomly initialized variables.

The Student's $t$-test analyzes the means of the sample populations generated by PSO and GA over the different runs, whereas ANOVA analyzes the variances. These statistical tools are used for determining whether the samples are generated from the same population, that is, the null hypothesis ($H_0$: means are equal) against the alternative hypothesis ($H_1$: at least one mean is different). M-ANOVA is used to compare the means of more than two populations with multiple dependent variables. The main idea of M-ANOVA is to examine the variances among means by evaluating the variation within populations, proportionate to the variations between populations.

The implementation of the fuzzy optimization methods have been performed on MATLAB and run on Intel(R) Xeon(R) processor with 16 GB of RAM (3.40 GHz, Windows 7, 64 bits).

### 6.1. Problem statements

The problem statements that we have considered for solving the two FMORRAP for $m$-state series-parallel and parallel-series systems are given as follows:

1) Series-Parallel FMORRAP

$$\text{Maximize} \quad \widetilde{F_s}(r,n) = \xi_1 \times \widetilde{\mathcal{R}_{s_1}}(\tilde{\mathcal{R}}_1, \tilde{\mathcal{R}}_2, \dots, \tilde{\mathcal{R}}_m) + \xi_2 \times \left(-\widetilde{C_s}\left(\tilde{C}_1, \tilde{C}_2, \dots, \tilde{C}_m\right)\right) \tag{37}$$

$$\text{subject to} \quad \sum_{i=1}^{m} w_i.(x_i + \exp(x_i/4)) \le W$$

$$\sum_{i=1}^{m} w_i v_i^2 n_i^2 \le V$$

$$\text{where} \quad r = (r_1, r_2, \dots, r_m);\ 0.5 \le r_i \le 1 - 10^6;\ n = (n_1, n_2, \dots, n_m);\ 1 \le n_i \le 5$$

$$\mathcal{R}_i(r,n) = 1 - \prod_{j=1}^{n_i}(1 - r_{i,j}):\ \mu_{\tilde{\mathcal{R}}_{s_1}}(x) = \left[\underline{\mu}_{\tilde{\mathcal{R}}_{s_1}}(x), \overline{\mu}_{\tilde{\mathcal{R}}_{s_1}}(x)\right];\ \ C_i(r,n) = \alpha_i\left(-\frac{t}{\ln(r_i)}\right)^{\beta_i}.(n_i + \exp(n_i/4));$$

$$\upsilon_{\tilde{C}_s}(x) = \left[\underline{\upsilon}_{\tilde{C}_s}(x), \overline{\upsilon}_{\tilde{C}_s}(x)\right];\ \mathcal{R}_i^L \le \mathcal{R}_i \le \mathcal{R}_i^R \quad C_i^L \le C_i \le C_i^R,\ i = 1,2,\dots,m.$$

2) Parallel-Series FMORRAP

$$\text{Maximize} \quad \widetilde{F_s}(r,n) = \xi_1 \times \widetilde{\mathcal{R}_{s_2}}(\tilde{\mathcal{R}}_1, \tilde{\mathcal{R}}_2, \dots, \tilde{\mathcal{R}}_m) + \xi_2 \times \left(-\widetilde{C_s}\left(\tilde{C}_1, \tilde{C}_2, \dots, \tilde{C}_m\right)\right) \tag{38}$$

$$\text{Subject to} \quad \sum_{i=1}^{m} w_i.(x_i + \exp(x_i/4)) \le W$$

$$\sum_{i=1}^{m} w_i v_i^2 n_i^2 \le V$$

$$\text{where} \quad r = (r_1, r_2, \dots, r_m);\ 0.5 \le r_i \le 1 - 10^6;\ n = (n_1, n_2, \dots, n_m);\ 1 \le n_i \le 5$$

$$\mathcal{R}_i(r,n) = 1 - \prod_{j=1}^{n_i} r_{i,j}\ ;\ \mu_{\tilde{\mathcal{R}}_{s_2}}(x) = \left[\underline{\mu}_{\tilde{\mathcal{R}}_{s_2}}(x), \overline{\mu}_{\tilde{\mathcal{R}}_{s_2}}(x)\right];\ \mathcal{R}_i^L \le \mathcal{R}_i \le \mathcal{R}_i^R;$$

$$C_i(r,n) = \alpha_i\left(-\frac{t}{\ln(r_i)}\right)^{\beta_i}.(n_i + \exp(n_i/4));\ \upsilon_{\tilde{C}_s}(r,n) = \left[\underline{\upsilon}_{\tilde{C}_s}(x), \overline{\upsilon}_{\tilde{C}_s}(x)\right];\quad C_i^L \le C_i \le C_i^R,\ i = 1,2,\dots,m.$$

In the two FMORRAPs, as given in Eq. (37) and Eq. (38), the $\widetilde{\mathcal{R}_{s_1}}$, $\widetilde{\mathcal{R}_{s_2}}$ and $\widetilde{C_s}$ are evaluated using Eq. (24), Eq. (27), and Eq. (29) respectively, where the $i$-th IT2 MFs of sub-system reliability ($\tilde{\mathcal{R}}_i$) and cost ($\tilde{C}_i$) are modeled with the IT2 TMFs as follows (discussed in Section 5.2):

$$\underline{\mu}_{\tilde{\mathcal{R}}_i}\left(x_i; \underline{\mathcal{R}}_i^L, \mathcal{R}_i, \underline{\mathcal{R}}_i^R\right) = \begin{cases} 0, & x_i \le \underline{\mathcal{R}}_i^L \\ \frac{x_i - \underline{\mathcal{R}}_i^L}{\mathcal{R}_i - \underline{\mathcal{R}}_i^L}, & \underline{\mathcal{R}}_i^L \le x_i \le \mathcal{R}_i \\ \frac{\underline{\mathcal{R}}_i^R - x_i}{\underline{\mathcal{R}}_i^R - \mathcal{R}_i}, & \mathcal{R}_i \le x_i \le \underline{\mathcal{R}}_i^R \\ 0, & x_i \le \underline{\mathcal{R}}_i^R \end{cases};\qquad \overline{\mu}_{\tilde{\mathcal{R}}_i}\left(x_i; \overline{\mathcal{R}}_i^L, \mathcal{R}_i, \overline{\mathcal{R}}_i^R\right) = \begin{cases} 0, & x_i \le \overline{\mathcal{R}}_i^L \\ \frac{x_i - \overline{\mathcal{R}}_i^L}{\mathcal{R}_i - \overline{\mathcal{R}}_i^L}, & \overline{\mathcal{R}}_i^L \le x_i \le \mathcal{R}_i \\ \frac{\overline{\mathcal{R}}_i^R - x_i}{\overline{\mathcal{R}}_i^R - \mathcal{R}_i}, & \mathcal{R}_i \le x_i \le \overline{\mathcal{R}}_i^R \\ 0, & x_i \le \overline{\mathcal{R}}_i^R \end{cases}$$

$$\overline{\upsilon}_{\tilde{C}_i}\left(x_i; \overline{C}_i^L, C_i, \overline{C}_i^R\right) = \begin{cases} 0, & x_i \le \overline{C}_i^L \\ \frac{x_i - \overline{C}_i^L}{C_i - \overline{C}_i^L}, & \overline{C}_i^L \le x_i \le C_i \\ \frac{\overline{C}_i^R - x_i}{\overline{C}_i^R - C_i}, & C_i \le x_i \le \overline{C}_i^R \\ 0, & x_i \le \overline{C}_i^R \end{cases};\qquad \underline{\upsilon}_{\tilde{C}_i}\left(x_i; \underline{C}_i^L, C_i, \underline{C}_i^R\right) = \begin{cases} 0, & x_i \le \underline{C}_i^L \\ \frac{x_i - \underline{C}_i^L}{C_i - \underline{C}_i^L}, & \underline{C}_i^L \le x_i \le C_i \\ \frac{\underline{C}_i^R - x_i}{\underline{C}_i^R - C_i}, & C_i \le x_i \le \underline{C}_i^R \\ 0, & x_i \le \underline{C}_i^R \end{cases}$$

### 6.2. Parameter settings

In the proposed FMORRAP, $r_i$ and $n_i$ are the decision variables which are randomly initialized as real numbers. However, before evaluating the objective functions, $n_i$ is transformed into the nearest integer. The common parameters of PSO and GA are chosen to be the same, i.e. population size (=100) and number of iterations (=100). The parameters of PSO, i.e. the acceleration coefficients $c_1$ and $c_2$ are chosen as $c_1 = c_2 = 1.5$ and $k$ is randomly selected from the uniform distribution as suggested in (Shi and Eberhart 1998; Beielstein 2002; Trelea 2003). The parameters of GA, i.e. crossover probability of one point crossover and mutation probability of one point mutation, are taken as 0.6 and 0.4, respectively. The tournament selection method is used to generate the reproduction.

We have considered five different weight vectors $[\xi_1, \xi_2] = \{[1,1], [1,0.5], [0.8,0.2], [0.2,0.8], [0.5,1]\}$ corresponding to the objectives. In the weight vector $[1,1]$ the expert insists on having no bias towards the maximizing reliability or minimizing cost. However, for the remaining weight vectors, we assume different importance to the objectives and assign varying weights to represent the biases towards the objectives. In other words, we explore both objectives, i.e. either the maximization of reliability or minimization of cost, in the FMORRAP models.

### 6.3. Result of series-parallel FMORRAP formulation

For the series-parallel FMORRAP formulation, the input dataset of a pharmaceutical plant (Garg and Sharma 2013), as given in Table 2, was used. In Table 2, the shaping factor $(\alpha)$, scaling factor $(\beta)$, component weights $(w)$ and volumes $(v)$ are provided for each redundant component of 10 different sub-systems used as the input parameters of the system. The upper limits of the total system cost $(C)$, weight $(W)$ and volume $(V)$ are also given, which provide the feasible region. The left and right ideal values of the sub-system reliability and cost corresponding to each weight vector are given in Table 3. These values are used in the construction of the IT2 MFs of the total IT2 fuzzy reliability and cost functions of the system.

**Table 2:** Input data: for series-parallel system (Garg and Sharma 2013)

| $i$ | $10^5 * \alpha_i$ | $\beta_i$ | $w_i$ | $v_i$ | $V$ | $C$ | $W$ | $T(h)$ |
|---|---|---|---|---|---|---|---|---|
| **1** | 0.611360 | 1.5 | 9.0 | 4.0 | | | | |
| **2** | 4.032464 | 1.5 | 7.0 | 5.0 | | | | |
| **3** | 3.578225 | 1.5 | 5.0 | 3.0 | | | | |
| **4** | 3.654303 | 1.5 | 9.0 | 2.0 | | | | |
| **5** | 1.163718 | 1.5 | 9.0 | 3.0 | | | | |
| **6** | 2.966955 | 1.5 | 10.0 | 4.0 | 289 | 553 | 483 | 1000 |
| **7** | 2.045865 | 1.5 | 6.0 | 1.0 | | | | |
| **8** | 2.649522 | 1.5 | 5.0 | 1.0 | | | | |
| **9** | 1.982908 | 1.5 | 8.0 | 4.0 | | | | |
| **10** | 3.516724 | 1.5 | 6.0 | 4.0 | | | | |

**Table 3:** Ideal value of sub-systems used in series-parallel system (Garg and Sharma 2013)

| | $[\xi_1, \xi_2] = [1,1]$ | $[\xi_1, \xi_2] = [1,0.5]$ | $[\xi_1, \xi_2] = [0.8,0.2]$ | $[\xi_1, \xi_2] = [0.2,0.8]$ | $[\xi_1, \xi_2] = [0.5,1]$ |
|---|---|---|---|---|---|
| ${R_i}^L$ | 0.6452566 | 0.6108074 | 0.6065163 | 0.6307274 | 0.6044854 |
| ${R_i}^R$ | 0.8199358 | 0.8471696 | 0.9227051 | 0.8144518 | 0.8250805 |
| ${C_i}^L$ | 185.607332 | 209.480891 | 199.697687 | 196.9306419 | 202.865356 |
| ${C_i}^R$ | 470.195913 | 510.048771 | 505.900821 | 432.2714613 | 440.986316 |

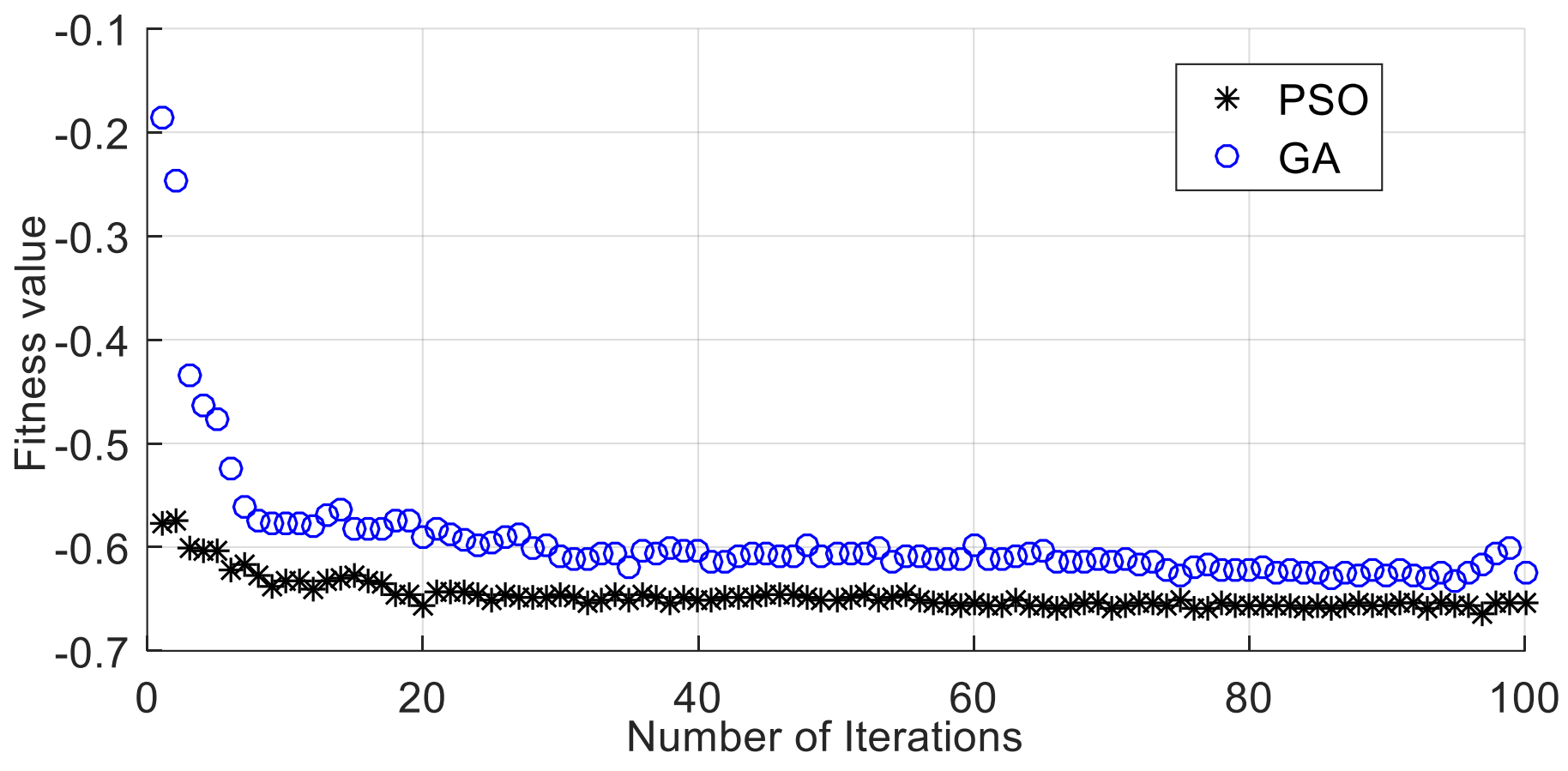

**Fig. 6:** Convergence of fitness function over the iterations

The IT2 fuzzy reliability and IT2 fuzzy cost functions of the FMORRAP conflict with each other; hence, the optimal results obtained will not be a single solution, but rather a set of solutions. These alternative solutions, having the properties of dominance, are called Pareto-optimal solutions (POSs) (Deb 2001). POSs allow the decision makers to choose a precise solution of the FMORRAP according to their preferences. The preferences, in the Pareto optimal region, are accommodated by the decision makers/system experts through assigning weights to each objective, i.e., maximization of reliability and minimization of cost. The comparison of convergence between PSO and GA for $[\xi_1, \xi_2] = [1,1]$ of series-parallel FMORRAP is shown in Fig. 6.

We have performed 50 different runs to obtain the best POF of the series-parallel configuration of the FMORRAP corresponding to the five different weight vectors (given in Table 3). The POF of the total system reliability and cost along with their respective component reliabilities, number of redundant components, the system weight and volume, are given in Table 4 and Table 5, solved using PSO and GA, respectively. We observed from the results obtained in Table 4 and Table 5 that aggregating the IT2 MFs of the sub-systems provides a definite preference (intermediate region of POF) among the POSs of FMORRAP with PSO whereas GA does not. This could be better explained as follows.

- The weight set, $[\xi_1, \xi_2] = [1.0,1.0]$, implies a POS set where the decision maker does not discriminate between the objectives, i.e., either to attain maximum reliability or minimum cost. This type of situation in selecting the Pareto optimal set of a particular POS is called a no-preference case.
- The weight sets, $[\xi_1, \xi_2] = [1.0,0.5]$ and $[\xi_1, \xi_2] = [0.5,1.0]$, imply that a particular Pareto optimal set is obtained where the decision maker has ranked the objectives according to the importance in the system, i.e., either to maximize reliability or minimize cost. These types of situations cause the occurrence of Pareto optimal sets of the extreme points of the Pareto optimal front.
- Similar to the previous case, the weight sets, $[\xi_1, \xi_2] = [0.8,0.2]$ and $[\xi_1, \xi_2] = [0.2,0.8]$, imply a particular Pareto optimal set is obtained where the decision maker has relatively ranked the objectives according to their importance. However, they do not occur in the extreme points of the POF, but somewhere in the intermediate region.

The best POF obtained from the PSO and GA corresponding to the series-parallel formulation of FMORRAP is depicted in Fig. 7 for the outputs in Tables 4-5. The above discussion regarding weight distribution and obtaining the preference based POF can be easily analyzed by Fig. 7 (a)-(b). From Fig. 7 (a), we observe that the POF has a good distribution and spread for the FMORRAP solved by PSO, while in Fig. 7 (b), the spread is good, but the distribution is bad for GA. In Fig. 7 (a), we observe that the weight set $[\xi_1, \xi_2] = [1.0,1.0]$ obtained the bottom left POS, $[\xi_1, \xi_2] = [0.5,1.0]$ obtained the top right POS, and $[\xi_1, \xi_2] = [0.8,0.2]$ obtained the intermediate POS of the POF. However, in Fig. 7(b), weight sets do not present as any of the extreme solutions, as $[\xi_1, \xi_2] = [1.0,0.5]$ and $[\xi_1, \xi_2] = [0.5,1.0]$, representing the maximized reliability and minimized cost respectively, make up the top right solutions of the POF. The weight sets $[\xi_1, \xi_2] = [1.0,1.0]$ and $[\xi_1, \xi_2] = [0.8,0.2]$, make up the bottom left solutions. However, the weight set $[\xi_1, \xi_2] = [0.2,0.8]$ is not depicted in the Fig. 7 (a)-(b) as the results obtained appears outside the range. Hence, from Tables 4-5 and Fig. 7, it can be seen that the POF obtained through PSO is comparatively better than GA in terms of preferences provided by the system experts towards the reliability and cost.

**Table 4:** Optimal results for series-parallel system using PSO

| $[\xi_1, \xi_2]$ | $r_i$ | $n_i$ | $\mathcal{R}_S$ | $C_S$ | $W_S$ | $V_S$ |
|---|---|---|---|---|---|---|
| **[1.0,1.0]** | (0.728966, 0.769185, 0.826255, 0.755018, 0.72388, 0.807148, 0.765235, 0.887747, 0.875649, 0.860357) | (3,3,3,3,3,3,3,2,2,2) | **0.867611877** | **437.0751367** | 411.956411 | 234 |
| **[1.0,0.5]** | (0.85, 0.792164, 0.875391, 0.770279, 0.912668, 0.825766, 0.713281, 0.752163, 0.862834, 0.731219) | (3,4,3,3,2,3,4,3,2,3) | **0.911134656** | **484.0057968** | 476.851180 | 286 |
| **[0.8,0.2]** | (0.850737, 0.864102, 0.723045, 0.637486, 0.941978, 0.80483, 0.800036, 0.855636, 0.833858, 0.7766) | (2,2,3,4,2,3,3,2,3,3) | **0.873018805** | **457.8556232** | 419.066424 | 228 |
| **[0.2,0.8]** | (0.7575, 0.760165, 0.687857, 0.697383, 0.643635, 0.785265, 0.956551, 0.693701, 0.781427, 0.662731) | (3,2,2,3,3,3,3,3,2,3) | **0.685369349** | **544.5856839** | 408.902853 | 219 |
| **[0.5,1.0]** | (0.904633, 0.827365, 0.783331, 0.85, 0.855563, 0.820486, 0.803125, 0.636795, 0.90646, 0.747079) | (2,3,3,3,3,3,3,4,2,3) | **0.916645936** | **485.9664204** | 440.674162 | 246 |

**Table 5:** Optimal results for series-parallel system using GA

| $[\xi_1, \xi_2]$ | $r_i$ | $n_i$ | $\mathcal{R}_S$ | $C_S$ | $W_S$ | $V_S$ |
|---|---|---|---|---|---|---|
| **[1.0,1.0]** | (0.728968755, 0.769197129,0.826289666, 0.755018985, 0.723910934, 0.8072603, 0.765263955, 0.887771541, 0.875658698, 0.860367419) | (3,3,3,3,3,2, 3,3,3,3) | **0.877430097** | **439.4384286** | 500.270007 | 259 |
| **[1.0,0.5]** | (0.85, 0.792163854, 0.875405107, 0.770346232, 0.912698009, 0.825775757, 0.713373152, 0.752163182, 0.86283449, 0.731218973) | (3,4,3,3,2,3, 4,3,2,3) | **0.911158623** | **484.0942288** | 476.851180 | 286 |
| **[0.8,0.2]** | (0.850005185, 0.792180169, 0.875393422, 0.770278993, 0.912668083, 0.825777087, 0.713320922, 0.752202532, 0.8628698, 0.731218973) | (3,4,3,3,2,3, 4,3,2,3) | **0.892344917** | **431.6762419** | 429.928502 | 236 |
| **[0.2,0.8]** | (0.757601057, 0.760233126, 0.687873854, 0.697422729, 0.64374898, 0.78532106, 0.956582946, 0.693738055, 0.781451985, 0.662731088) | (3,2,2,3,3,3, 3,3,2,3) | **0.685472781** | **545.0438887** | 408.902853 | 219 |
| **[0.5,1.0]** | (0.904695338, 0.827385222, 0.783441183, 0.85, 0.85564716, 0.820648517, 0.803255173, 0.636835557, 0.906459969, 0.747085507) | (2,3,3,3,3,3, 3,4,2,3) | **0.916714694** | **486.1984923** | 440.674162 | 246 |

As discussed above, the PSO based solution approach found the optimal or nearly optimal Pareto solutions to the FMORRAP across all five weight vectors. To further examine the performances of the algorithms, we conduct a statistical comparison between PSO and GA for series-parallel FMORRAP. The statistical comparisons include mean, median, standard deviation, $t$-test and multivariate analysis of variance (M-ANOVA) for the 50 replications of each algorithm over five weight vectors, and the results are given in Table 6.

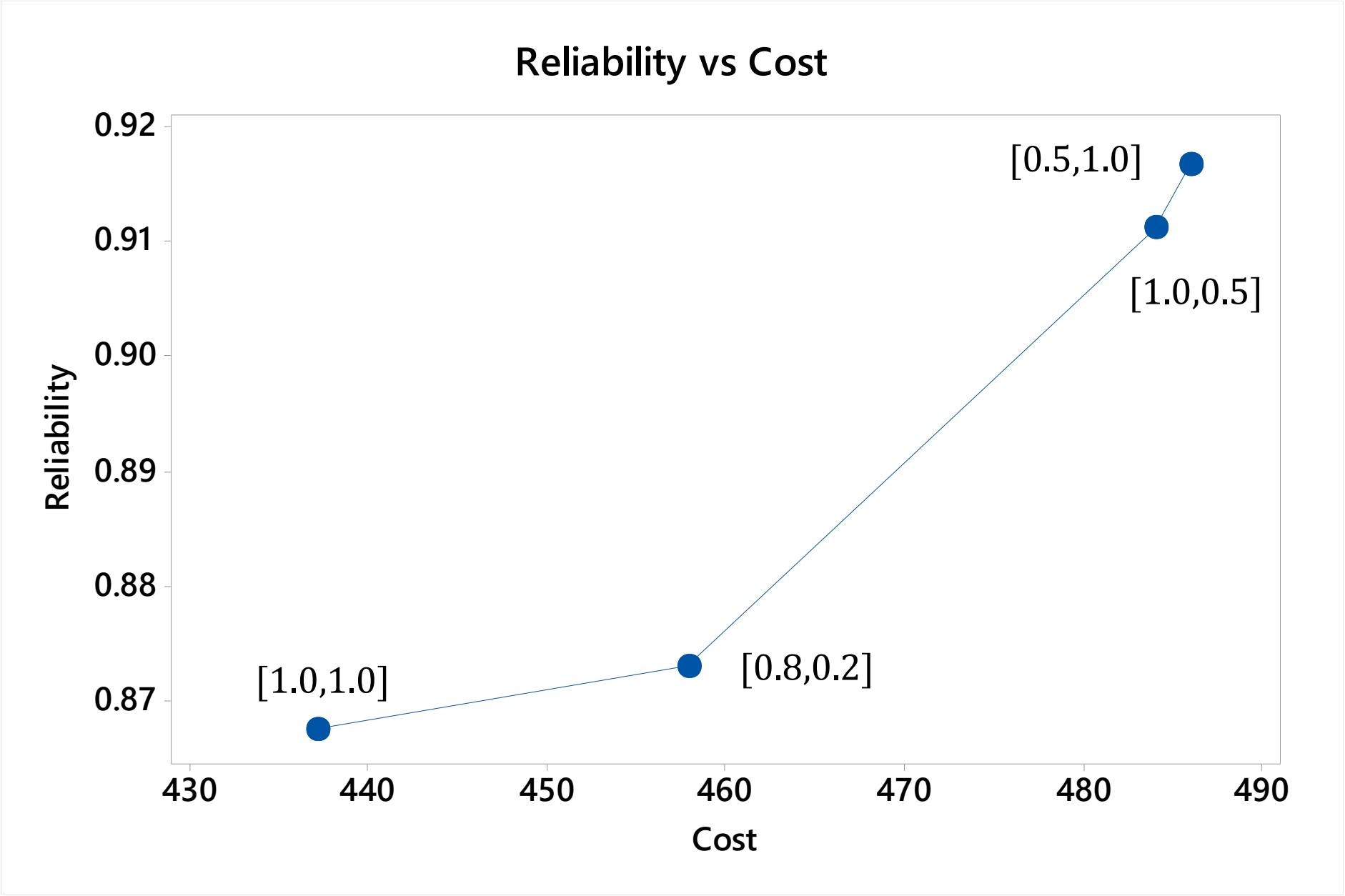

(a)

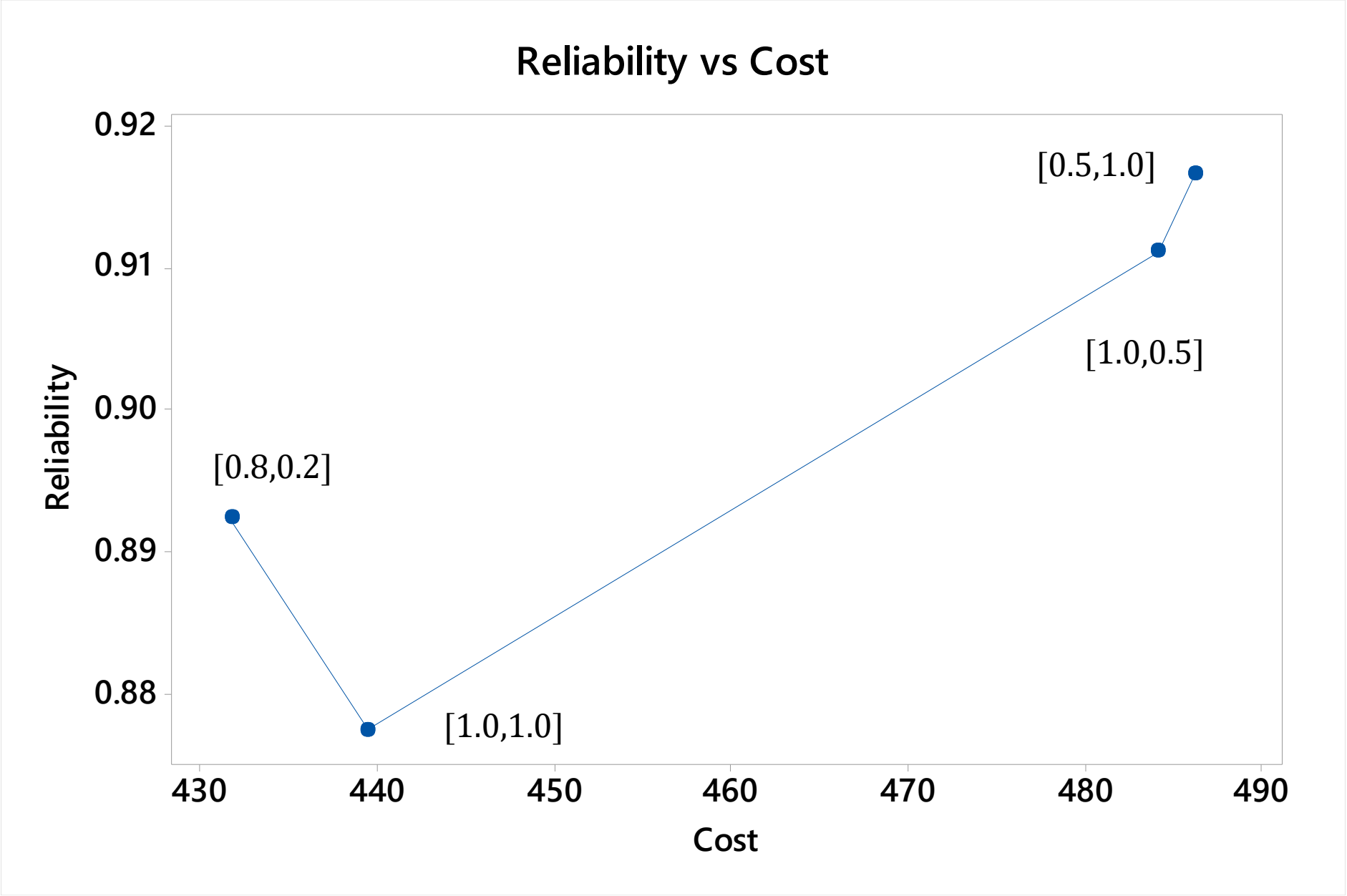

(b)

**Fig. 7:** Pareto-optimal front for series-parallel FMORRAP formulation using (a) PSO, and (b) GA

From Table 6, there is a clear-cut statistical difference among the reliability and cost obtained through PSO and GA, corresponding to each weight vector, as mean, median and SD of the objective function values are different. Moreover, the PSO outperformed the GA algorithm on the selected weights, that is, the average SDs obtained by PSO are much smaller than those of GA. Further, to verify that PSO and GA algorithms have significant difference in their performance, we conducted $t$-test and M-ANOVA. The $t$-test examines whether the population means of two solutions vary from one another, whereas M-ANOVA examines the population variances. Both of these tests showed that algorithms are significant corresponding to the hypotheses. Since the FMORRAP generates the normal distribution for each objective, the two significant factors, $t$-value and $p$-value (for t-test) and $f$-value and $p$-value (for M-ANOVA) are depicted in Table 6. The $t$-test results showed that the $p$-values are significant, that is, less than $0.005$ for both PSO and GA algorithms, however, $t-$values are higher for PSO than for GA. In other words, PSO outperformed GA on the weight sets in the statistical $t$-test. In the same manner, the M-ANOVA results are significant with $p-\text{value} < 0.005$ and $f-\text{value} > 0$ or higher, that is, PSO outperformed GA on the weight sets in the statistical sense.

**Table 6:** Statistical analysis: comparison between PSO and GA for series-parallel system

| Parameters | | $[\mathbf{1.0}, \mathbf{1.0}]$ | | $[\mathbf{0.5}, \mathbf{1.0}]$ | | $[\mathbf{1.0}, \mathbf{0.5}]$ | | $[\mathbf{0.8}, \mathbf{0.2}]$ | | $[\mathbf{0.2}, \mathbf{0.8}]$ | |
|---|---|---|---|---|---|---|---|---|---|---|---|
| | | **PSO** | **GA** | **PSO** | **GA** | **PSO** | **GA** | **PSO** | **GA** | **PSO** | **GA** |
| Samples | $N$ | 50 | 50 | 50 | 50 | 50 | 50 | 50 | 50 | 50 | 50 |
| Mean | $(\mathcal{R}_S, C_S)$ | (0.890, 478.539) | (0.674, 374.931) | (0.913, 456.905) | (0.746, 402.790) | (0.908, 480.763) | (0.823, 450.793) | (0.867, 439.076) | (0.879, 484.394) | (0.676, 471.138) | (0.531, 307.771) |
| SD | $(\mathcal{R}_S, C_S)$ | (0.006, 17.302) | (0.351, 160.283) | (0.003, 5.565) | (0.259, 129.112) | (0.019, 11.273) | (0.206, 107.608) | (9.94E-06, 0.025) | (0.127, 73.089) | (1.2E-05,0.150) | (0.299, 136.379) |
| Median | $(\mathcal{R}_S, C_S)$ | (0.889, 474.685) | (0.861, 438.324) | (0.914, 455.079) | (0.842, 444.234) | (0.916, 486.070) | (0.881, 478.080) | (0.866, 439.072) | (0.897, 495.003) | (0.675, 471.141) | (0.660, 327.844) |
| $t$-*Test* | $t$-value | 195.21 | 16.51 | 579.39 | 22.02 | 300.99 | 29.57 | 121979.21 | 46.78 | 22152.4 | 15.93 |
| | $p$-value | 0.000 | 0.000 | 0.000 | 0.000 | 0.000 | 0.000 | 0.000 | 0.000 | 0.000 | 0.000 |
| M-ANOVA | $f$-value | 20.65 | | 8.77 | | 3.84 | | 19.22 | | 71.75 | |
| | $p$-value | 0.000 | | 0.004 | | 0.0523 | | 0.000 | | 0.000 | |

### *6.4. Result of parallel-series FMORRAP formulation*

For solving the parallel-series FMORRAP formulation, the input dataset (Mutingi 2014) as shown in Table 7, where the shaping factor $(\alpha)$, scaling factor $(\beta)$, component weights $(w)$ and volumes $(v)$ are provided for each redundant component of 5 different sub-systems used as the input parameters of the system. The upper limits of the total system cost $(C)$, weight $(W)$ and volume $(V)$ are also given, which help form the feasible region. The sub-system reliability and cost left and right ideal values corresponding to each of the weight vectors are given in Table 8. These values were used in the creation of the IT2 MFs of the total IT2 fuzzy reliability and cost functions of the system.

Similar to the series-parallel configuration of the FMORRAP, we have performed the 50 different runs of the PSO and GA algorithms to obtain the best POF of parallel-series FMORRAP corresponding to the five different weight vectors (given in Table 8). The POF of the total system reliability and cost along with their corresponding component reliabilities, number of redundant components, system weight and volume are given in Table 9 and Table 10, solved using PSO and GA, respectively. From Tables 9-10, we can observe that the optimal solutions obtained by aggregating the IT2 MFs of the sub-systems provide the intermediate region of the POF.

**Table 7:** Input data: for parallel-series system (Mutingi 2014)

| $i$ | $10^5\alpha_i$ | $\beta_i$ | $w_i v_i^2$ | $w_i$ | $V$ | $C$ | $W$ |
|---|---|---|---|---|---|---|---|
| **1** | 2.330 | 1.5 | 1 | 7 | | | |
| **2** | 1.450 | 1.5 | 2 | 8 | | | |
| **3** | 0.541 | 1.5 | 3 | 8 | 110 | 175 | 200 |
| **4** | 8.050 | 1.5 | 4 | 6 | | | |
| **5** | 1.950 | 1.5 | 2 | 9 | | | |

**Table 8**: The ideal value of sub-systems for parallel-series system (Mutingi 2014)

| | $[\xi_1,\xi_2]=[1,1]$ | $[\xi_1,\xi_2]=[1,0.5]$ | $[\xi_1,\xi_2]=[0.8,0.2]$ | $[\xi_1,\xi_2]=[0.2,0.8]$ | $[\xi_1,\xi_2]=[0.5,1]$ |
|---|---|---|---|---|---|
| ${R_i}^L$ | 0.6 | 0.6 | 0.6 | 0.6 | 0.6 |
| ${R_i}^R$ | 0.9999 | 0.9999 | 0.9999 | 0.9999 | 0.9999 |
| ${C_i}^L$ | 60 | 60 | 60 | 60 | 60 |
| ${C_i}^R$ | 180 | 180 | 180 | 180 | 180 |

**Table 9:** Optimal results for parallel-series system using PSO

| $[\xi_1,\xi_2]$ | $r_i$ | $n_i$ | $\mathcal{R}_S$ | $C_S$ | $W_S$ | $V_S$ |
|---|---|---|---|---|---|---|
| **[1.0,1.0]** | (0.583327,0.697626,0.562849,0.698137, 0.553977) | (3,2,2,4,2) | 0.851456 | 103.0558 | 192.1318 | 101 |
| **[1.0,0.5]** | (0.530894,0.590572,0.627185,0.698474, 0.65294) | (2,3,2,4,2) | 0.848787 | 101.0737 | 195.1854 | 106 |
| **[0.8,0.2]** | (0.524414,0.7,0.7,0.515625,0.53125) | (3,2,2,3,3) | 0.836724 | 48.49055 | 192.4811 | 83 |
| **[0.2,0.8]** | (0.616663,0.539099,0.611383,0.699818,0.558962, 0.827283) | (2,2,2,4, 3) | 0.827283 | 100.56392 | 198.2389 | 106 |
| **[0.5,1.0]** | (0.543701,0.654936,0.672852,0.7, 0.588135) | (2,2,2,4,3) | 0.866748 | 102.6775 | 198.2389 | 106 |

**Table 10**: Optimal results for parallel-series system using GA

| $[\xi_1,\xi_2]$ | $r_i$ | $n_i$ | $\mathcal{R}_S$ | $C_S$ | $W_S$ | $V_S$ |
|---|---|---|---|---|---|---|
| **[1.0,1.0]** | (0.583396, 0.697682,0.562926,0.698139, 0.554059) | (3,2,2,4,2) | 0.85153 | 103.0647 | 192.1318 | 101 |
| **[1.0,0.5]** | (0.530988,0.59066,0.627247,0.698474, 0.65294) | (2,3,2,4,2) | 0.848845 | 101.0792 | 195.1854 | 106 |
| **[0.8,0.2]** | (0.530984,0.590674,0.627293,0.698479, 0.65296) | (3,2,2,3,3) | 0.840269 | 85.76012 | 192.4811 | 83 |
| **[0.2,0.8]** | (0.530152,0.6215841,0.580421,0.6982704, 0.603683) | (2,3,2,4,2) | 0.824504 | 99.7639 | 195.1854 | 106 |
| **[0.5,1.0]** | (0.531033,0.590695,0.627281,0.698489, 0.652993) | (3,2,2,4,2) | 0.853253 | 101.477 | 192.1318 | 101 |

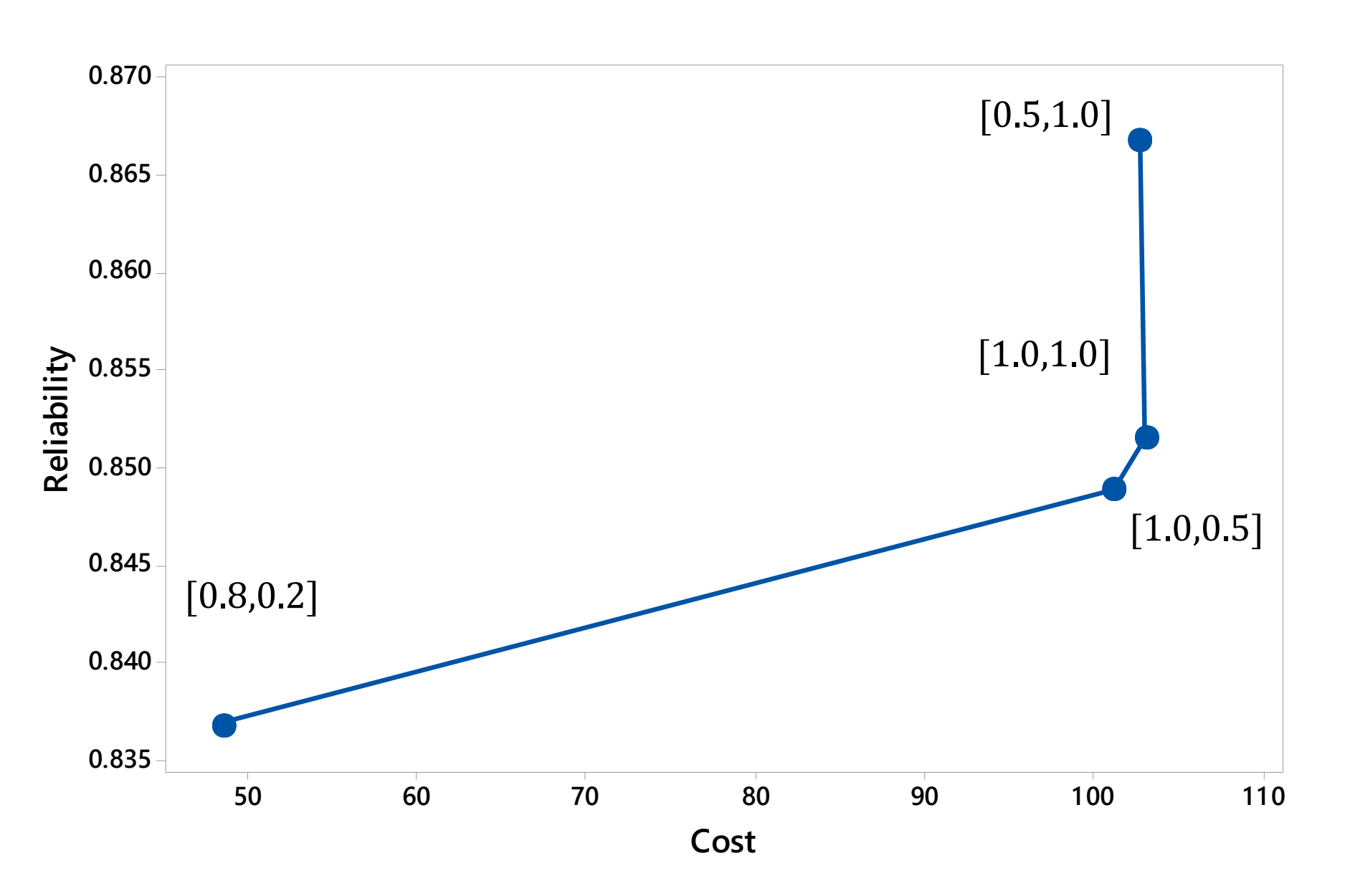

(a)

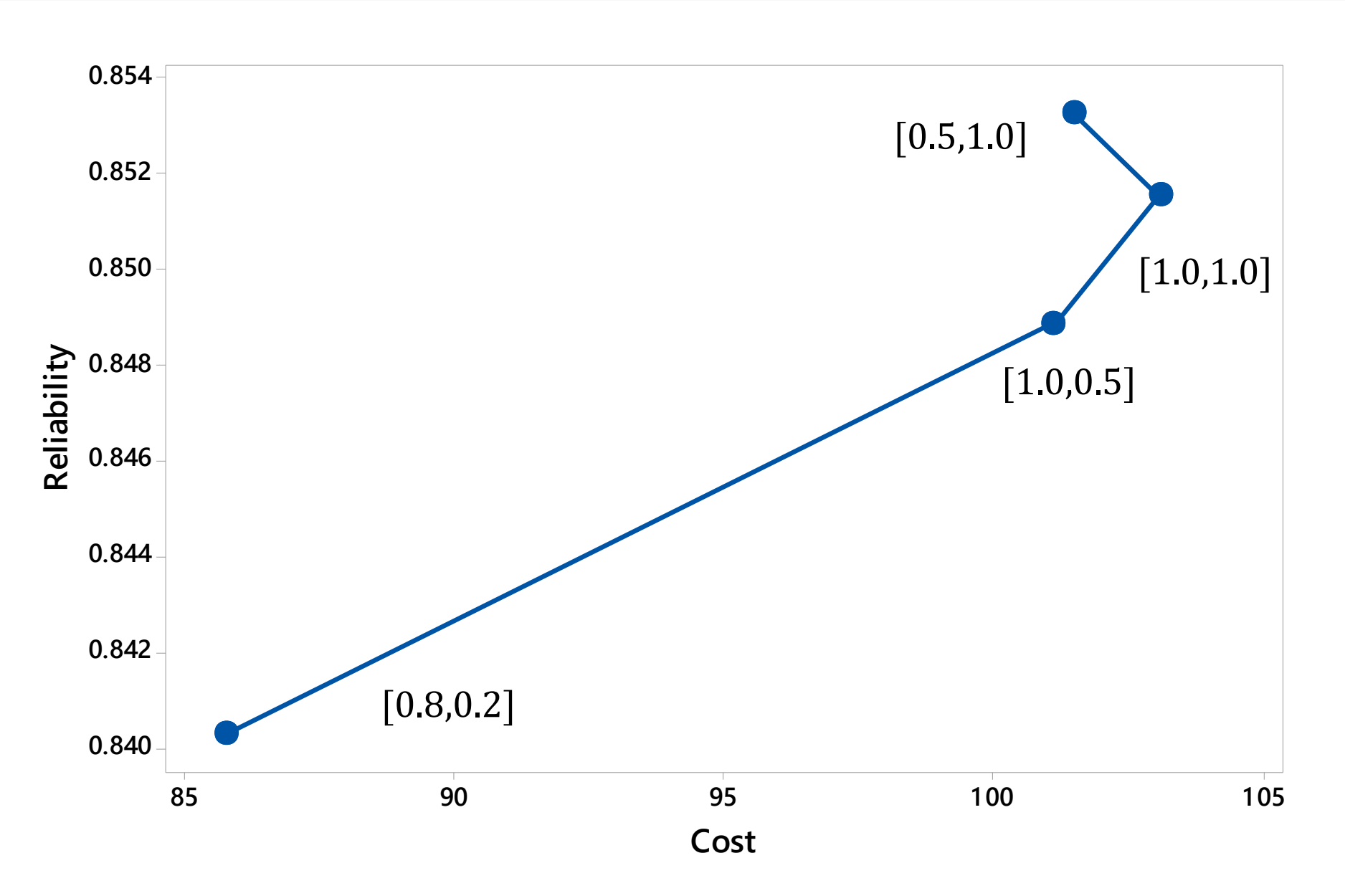

(b)

**Fig. 8:** Pareto-optimal front for parallel-series FMORRAP formulation using (a) PSO, and (b) GA

The best POF obtained from the PSO and GA corresponding to the parallel-series formulation of FMORRAP is depicted in Fig. 8 for the outputs in Tables 9-10. From Fig. 8 (a), we observe that the POF has a good distribution and spread for the FMORRAP solved by PSO, while in Fig. 8 (b), the spread is good, but the distribution is bad for GA. Being aware of the weight vectors, the preference based POF can be easily analyzed by Fig. 8 (a)-(b). In Fig. 8 (a)-(b), we observe that the weight set $[\xi_1, \xi_2] = [0.8,0.2]$ obtained the bottom-left Pareto optimal solution for PSO and GA, representing the situation when the decision maker has relatively ranked the objectives according to their importance. The weight set $[\xi_1, \xi_2] = [0.5,1.0]$ and $[\xi_1, \xi_2] = [1.0,0.5]$ obtained the top-right and intermediate Pareto optimal solution, respectively, for PSO and GA. They imply those Pareto optimal sets that are chosen by the decision maker according to the rank of objectives i.e., maximize reliability or minimize cost. The weight set $[\xi_1, \xi_2] = [1.0,1.0]$, called a no-preference case, implies a Pareto optimal set where the decision maker does not discriminate between the objectives. It obtains the perfect intermediate Pareto-optimal solution for PSO compared to GA, where it obtains the intermediate solution towards the top-right of the POF. However, the weight set $[\xi_1, \xi_2] = [0.2,0.8]$ is not depicted in the Fig. 8 (a)-(b) as the results obtained appears outside the range. Hence, from the above discussion, we can say that the POF obtained through PSO in parallel-series configuration of FMORRAP is comparatively better than GA with respect to each weight vector given by the system experts towards reliability and cost.

Similar to the tests conducted in the previous sub-section, we conducted a statistical comparison between PSO and GA for parallel-series FMORRAP. The statistical comparisons include mean, median, standard deviation, $t$-test and multivariate analysis of variance (M-ANOVA) for the 50 replications of each algorithm over five weight vectors, the results for which are displayed in Table 11. From Table 11, a definite statistical difference among the reliability and cost obtained through PSO and GA is found corresponding to each weight vector, as mean, median and SD of the objective function values all differ. Moreover, the PSO outperformed the GA algorithm on the selected weights, i.e, the average SDs obtained by PSO are much smaller than those of GA. Further, to verify that PSO and GA algorithms have significant difference in their performance, we conducted $t$-test and M-ANOVA. Given that the FMORRAP generates a normal distribution for the objectives, the two significant factors, $t$-value and $p$-value (for t-test) and $f$-value and $p$-value (for M-ANOVA) are depicted in Table 11. The $t$-test results show significant $p$-values, that is, $p - \text{value} < 0.005$ for both PSO and GA algorithms, and show that the $t$-values are higher for PSO as compared to GA, concluding that PSO outperformed GA on the weight sets. Similarly, the M-ANOVA results have significant with $p$-value, i.e., less than $0.005$, and $f - \text{value} > 0$ or higher, showing that PSO outperformed GA on the weight sets in the statistical sense.

It is to be noticed here that, for our proposed FMORRAP, the PSO based solution approach is superior to GA. However, we cannot assert that it will be superior for all problem sets, as there is no guarantee that one optimization algorithm that is effective on one set of problems will be as effective on another set of problems, proved by Wolpert and Macready's *no free lunch theorems* in (Wolpert and Macready 1997).

**Table 11:** Statistical analysis: comparison between PSO and GA for parallel-series system

| Parameters | | $[\mathbf{1.0}, \mathbf{1.0}]$ | | $[\mathbf{0.5}, \mathbf{1.0}]$ | | $[\mathbf{1.0}, \mathbf{0.5}]$ | | $[\mathbf{0.8}, \mathbf{0.2}]$ | | $[\mathbf{0.2}, \mathbf{0.8}]$ | |
|---|---|---|---|---|---|---|---|---|---|---|---|
| | | **PSO** | **GA** | **PSO** | **GA** | **PSO** | **GA** | **PSO** | **GA** | **PSO** | **GA** |
| Samples | $N$ | 50 | 50 | 50 | 50 | 50 | 50 | 50 | 50 | 50 | 50 |
| Mean | $(\mathcal{R}_S, C_S)$ | (0.844, 98.743) | (0.806, 99.256) | (0.864, 102.401) | (0.838, 100.402) | (0.849, 101.094) | (0.797, 97.923) | (0.836, 48.299) | (0.785, 90.600) | (0.858, 100.387) | (0.845, 100.286) |
| SD | $(\mathcal{R}_S, C_S)$ | (2.77E-5,0.010) | (0.023, 1.614) | (2.4E-5, 0.0119) | (0.025, 2.673) | (3.16E-5,0.011) | (0.020, 3.463) | (4.07E-5,0.015) | (0.017, 10.838) | (0.0114, 2.778) | (0.029, 2.875) |
| Median | $(\mathcal{R}_S, C_S)$ | (0.889, 474.685) | (0.861, 438.324) | (0.914, 455.079) | (0.842, 444.234) | (0.916, 486.070) | (0.881, 478.080) | (0.866, 439.072) | (0.897, 495.003) | (0.675, 471.141) | (0.660, 327.844) |
| $t$-Test | $t$-value | 68486.7 | 431.26 | 60017.8 | 263.41 | 67156.57 | 198.31 | 22901.6 | 58.60 | 253.38 | 244.57 |
| | $p$-value | 0.000 | 0.000 | 0.000 | 0.000 | 0.000 | 0.000 | 0.000 | 0.000 | 0.000 | 0.000 |
| M-ANOVA | $f$-value | 5.04 | | 27.96 | | 41.91 | | 761.72 | | 0.03 | |
| | $p$-value | 0.027 | | 0.000 | | 0.000 | | 0.000 | | 0.859 | |

## 7. Conclusion

The present paper solved the fuzzy multi-objective reliability-redundancy allocation problem (FMORRAP) of series-parallel and parallel-series systems using particle swarm optimization (PSO) and genetic algorithm (GA). The two objective functions of the problem, the maximization of reliability and the minimization of cost, have been designed under type-2 fuzzy uncertain environment. To do so, the reliability and cost of the sub-systems have been modeled with interval type-2 fuzzy membership functions (IT2 MFs) and the total tradeoff objective functions of the systems are evaluated based on their configuration using extension principle. The IT2 MFs accommodate the parametric, manufacturing environmental and designers' uncertainties associated with the system by capturing the multiple opinions from several system experts. The formulated optimization models are solved for suitable datasets and the results are presented in the form of a Pareto-optimal front. The Pareto-optimal solutions found by our proposed PSO based solution approach are all better than the solutions obtained by GA under the different weight vectors. To verify the significance of the FMORRAP between PSO and GA algorithms, we have conducted different runs and performed statistical analysis, namely mean, standard deviation, median, $t$-test and multivariate analysis of variance (M-ANOVA).

In the future work, the type-2 fuzzy multi-objective reliability-redundancy allocation problem could be formulated for more system configurations such as bridge, complex, $\boldsymbol{k}$-out-of-$\boldsymbol{n}$ systems and so on, and could be used to solve more real life examples such as water resource management (Dolatshahi-Zand and Khalili-Damghani 2015), paper plants (Komal and Sharma 2014), etc. Further, some recent and more efficient evolutionary approaches, such as parallel genetic colony (Chen and Chien 2011), parallel cat swarm (Tsai et al. 2008), enhanced parallel cat swarm optimization algorithms (Tsai et al. 2012) and so on, could be used to solve the FMORRAP.

**Acknowledgements**

The authors would like to express their sincere thanks to the reviewers, the associate editor, and the EIC of the journal for their helpful comments and suggestions that helped in the improvement of the manuscript. Authors gratefully acknowledge the infrastructural and research facilities provided by the South Asian University, New Delhi through the Computational Intelligence lab of the Department of Computer Science while designing the experiments and conducting investigations.